\documentclass[10pt]{article} %
\usepackage[accepted]{tmlr}

\usepackage{hyperref}
\usepackage{url}

\usepackage{colortbl}
\usepackage{xcolor}         %
\usepackage[utf8]{inputenc} %
\usepackage[T1]{fontenc}    %
\usepackage{url}            %
\usepackage{booktabs}       %
\usepackage{amsfonts}       %
\usepackage{nicefrac}       %
\usepackage{microtype}      %
\definecolor{mydarkblue}{rgb}{0,0.08,0.45}
\definecolor{myblue}{HTML}{3b75c3}
\definecolor{myred}{HTML}{E33222}
\definecolor{mygreen}{HTML}{438773}
\definecolor{mymaroon}{RGB}{142,27,19}
\definecolor{maroon}{HTML}{800000}
\definecolor{mycite}{cmyk}{0.55,1,0,0.15}
\definecolor{codeblue}{rgb}{0.25,0.5,0.5}
\definecolor{codekw}{rgb}{0.85, 0.18, 0.50}
\definecolor{codegreen}{rgb}{0,0.6,0}
\definecolor{codegray}{rgb}{0.5,0.5,0.5}
\definecolor{codepurple}{rgb}{0.58,0,0.82}
\definecolor{backcolour}{rgb}{0.95,0.95,0.92}
\definecolor{mygray}{gray}{0.925}
\definecolor{myGreen}{rgb}{0.261,0.460,0.343}
\definecolor{myRed}{rgb}{0.722,0.304,0.320}
       
\hypersetup{
    colorlinks=true,
    citecolor=mycite,
    urlcolor=mydarkblue,
    linkcolor = mydarkblue
}
\usepackage{algorithm}
\usepackage{algorithmic}
\usepackage{amsthm}
\usepackage{amsmath}
\usepackage{graphicx}
\usepackage{enumitem}
\usepackage{wrapfig}
\usepackage{bbm}
\usepackage{amsmath}
\usepackage{mathtools}
\usepackage{amsfonts}
\usepackage{bm}
\usepackage{booktabs}
\usepackage{bbm}
\usepackage{multirow}
\usepackage{enumitem}
\usepackage{wrapfig}
\usepackage[tableposition=bottom]{caption}
\usepackage{subcaption}
\usepackage{float}
\usepackage{mathrsfs}
\usepackage{color}
\usepackage{arydshln}
\usepackage{mathtools}
\usepackage{amsthm}
\usepackage{sansmath}
\usepackage{mleftright}
\usepackage{tikz}
\usepackage{footnote}
\makesavenoteenv{tabular}
\makesavenoteenv{table}
\usepackage{enumitem}
\usepackage{bbm}
\usepackage{pifont}

\usepackage{circledsteps}
\pgfkeys{/csteps/inner color=white}
\pgfkeys{/csteps/outer color=white}
\pgfkeys{/csteps/fill color=black}
\pgfkeys{/csteps/inner ysep=1pt}
\pgfkeys{/csteps/inner xsep=3pt}

\usepackage[capitalize,noabbrev,nameinlink]{cleveref}
\usepackage[textsize=tiny]{todonotes}

\setenumerate[1]{itemsep=0pt,partopsep=0pt,parsep=\parskip,topsep=5pt}
\setitemize[1]{itemsep=0pt,partopsep=0pt,parsep=\parskip,topsep=5pt}
\setdescription{itemsep=0pt,partopsep=0pt,parsep=\parskip,topsep=5pt}

\usepackage{blindtext}
\usepackage{titlesec}
\usepackage{multirow}
\usepackage{tabularray}
\usepackage[most]{tcolorbox}

\definecolor{mydarkblue}{rgb}{0,0.08,0.45}
\definecolor{myblue}{HTML}{3b75c3}
\definecolor{myred}{HTML}{E33222}
\definecolor{mygreen}{HTML}{438773}
\definecolor{mymaroon}{RGB}{142,27,19}
\definecolor{maroon}{HTML}{800000}
\definecolor{mycite}{cmyk}{0.55,1,0,0.15}
\definecolor{codeblue}{rgb}{0.25,0.5,0.5}
\definecolor{codekw}{rgb}{0.85, 0.18, 0.50}
\definecolor{codegreen}{rgb}{0,0.6,0}
\definecolor{codegray}{rgb}{0.5,0.5,0.5}
\definecolor{codepurple}{rgb}{0.58,0,0.82}
\definecolor{backcolour}{rgb}{0.95,0.95,0.92}
\definecolor{mygray}{gray}{0.925}
\definecolor{myred}{HTML}{E33222}
\newcommand{\ms}[2]{{$#1$\tiny{$\pm#2$}}}

\newcommand{\bluenode}{\vcenter{\hbox{\includegraphics[scale=0.39]{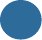}}}}
\newcommand{\greennode}{\vcenter{\hbox{\includegraphics[scale=0.39]{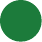}}}}

\newcommand{\orangenode}{\vcenter{\hbox{\includegraphics[scale=0.39]{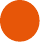}}}}
\newcommand{\bluelabel}{\vcenter{\hbox{\includegraphics[scale=0.75]{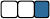}}}}
\newcommand{\dblegend}{\vcenter{\hbox{\includegraphics[scale=0.08]{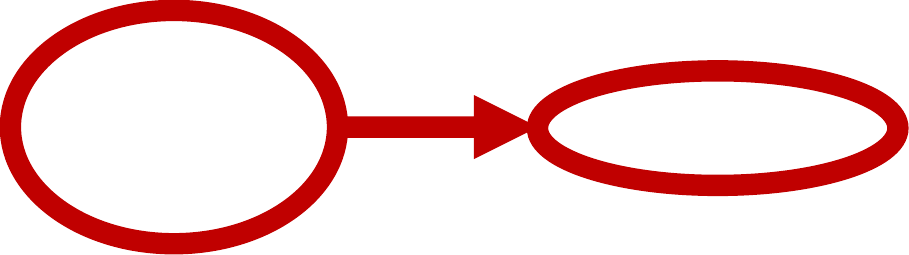}}}}

\newcommand{\cora}{\texttt{Cora}\xspace}
\newcommand{\coraml}{\texttt{FullCoraML}\xspace}
\newcommand{\citeseer}{\texttt{CiteSeer}\xspace}
\newcommand{\pubmed}{\texttt{PubMed}\xspace}

\newcommand{\hmlp}{{\fontsize{9}{11}\selectfont \textsf{HMLP}}\xspace}
\newcommand{\tmlp}{{\fontsize{9}{11}\selectfont \textsf{TMLP}}\xspace}

\newtcolorbox{applebox}[1]{
  enhanced,
  colback=white,
  colframe=gray!40,
  fonttitle=\bfseries,
  coltitle=black,
  colbacktitle=white,
  left=4pt,
  right=4pt,
  top=2pt,
  bottom=2pt,
  boxsep=4pt,
  boxrule=1.5pt,
  arc=1.0mm,
  title=#1, %
  borderline={0pt}{0pt}{white},
  attach boxed title to top left={yshift=-2mm, xshift=3mm},
  boxed title style={sharp corners, boxrule=0pt, colback=white, frame hidden, left=2pt, right=2pt},
}

\usepackage[loose]{minitoc}

\doparttoc %
\faketableofcontents %

\title{On the Equivalence of Graph Convolution and Mixup}

\author{\name Xiaotian Han$^{1}$\thanks{Corresponds to xhan@case.edu. The code is at \url{https://github.com/ahxt/GraphConv_is_Mixup}.}~~Hanqing Zeng$^{2}$~Yu Chen$^{3}$~Shaoliang Nie$^{2}$~ Jingzhou Liu$^{2}$~Kanika Narang$^{2}$~\\Zahra Shakeri$^{2}$~~Karthik Abinav Sankararaman$^{2}$~Song Jiang$^{4}$~Madian Khabsa$^{2}$~\\Qifan Wang$^{2}$~Xia Hu$^{5}$\vspace{2pt}\\
\vspace{4pt}
\addr $^{1}$Case Western Reserve University~~$^{2}$Meta AI~~$^{3}$Anytime AI~~$^{4}$UCLA~~$^{5}$Rice University 
}

\def\month{09}  %
\def\year{2024} %
\def\openreview{\url{https://openreview.net/forum?id=koC6zyaj73}} %

\begin{document}

\maketitle

\begin{abstract}
This paper investigates the relationship between graph convolution and Mixup techniques. Graph convolution in a graph neural network involves aggregating features from neighboring samples to learn representative features for a specific node or sample. On the other hand, Mixup is a data augmentation technique that generates new examples by averaging features and one-hot labels from multiple samples. One commonality between these techniques is their utilization of information from multiple samples to derive feature representation. This study aims to explore whether a connection exists between the two. Our investigation reveals that, under two mild modifications, graph convolution can be viewed as a specialized form of Mixup that is applied during both the training and testing phases. The two modifications are 1) \textit{Homophily Relabel} - assigning the target node's label to all its neighbors, and 2) \textit{Test-Time Mixup} - Mixup the feature during the test time. We establish this equivalence mathematically by demonstrating that graph convolution networks and simplified graph convolution can be expressed as a form of Mixup. We also empirically verify the equivalence by training an MLP using the two modifications to achieve comparable performance. 
\end{abstract}

\section{Introduction}\label{sec:intro}
Graph Neural Networks (GNNs)~\citep{wu2020comprehensive,zhou2020graph} have recently been recognized as the \textit{de facto} state-of-the-art algorithm for graph learning. The core idea behind GNNs is neighbor aggregation, which involves combining the features of a node's neighbors. Specifically, for a target node with feature $\mathbf{x}_i$, one-hot label $\mathbf{y}_i$, and neighbor set $\mathcal{N}_i$, the graph convolution operation in GCN is essentially as follows:
\begin{equation}\label{eq:gcn}
    (\Tilde{\mathbf{x}}, \Tilde{\mathbf{y}}) = \left(\frac{1}{|\mathcal{N}_i|} \sum_{k \in \mathcal{N}_i} \mathbf{x}_k,~~\mathbf{y}_i\right),
\end{equation} 
In parallel, Mixup~\citep{zhang2018mixup} is proposed to train deep neural networks effectively, which also essentially generates a new sample by averaging the features and labels of multiple samples:
\begin{equation}\label{eq:mixup}
(\Tilde{\mathbf{x}}, \Tilde{\mathbf{y}}) = \left(\sum_{i=1}^{N} \lambda_i\mathbf{x}_i,~~
\sum_{i=1}^{N} \lambda_i\mathbf{y}_i\right), \quad\text{s.t.}\quad \sum^{N}_{i=1} \lambda_i= 1,
\end{equation}
where $\mathbf{x}_i$/$\mathbf{y}_i$ are the feature/label of sample $i$. Mixup typically takes two data samples ($N=2$).

\cref{eq:gcn} and \cref{eq:mixup} highlight a remarkable similarity between graph convolution and Mixup, i.e., \textit{the manipulation of data samples through averaging the features}. This similarity prompts us to investigate the relationship between these two techniques as follows:
\begin{center}
    \vspace{-5pt}
    \emph{\textbf{Is there a connection between graph convolution and Mixup?}}
    \vspace{-5pt}
\end{center}

\begin{figure}[!t]
    \centering
    \vspace{-30pt}
    \includegraphics[width=0.9\textwidth]{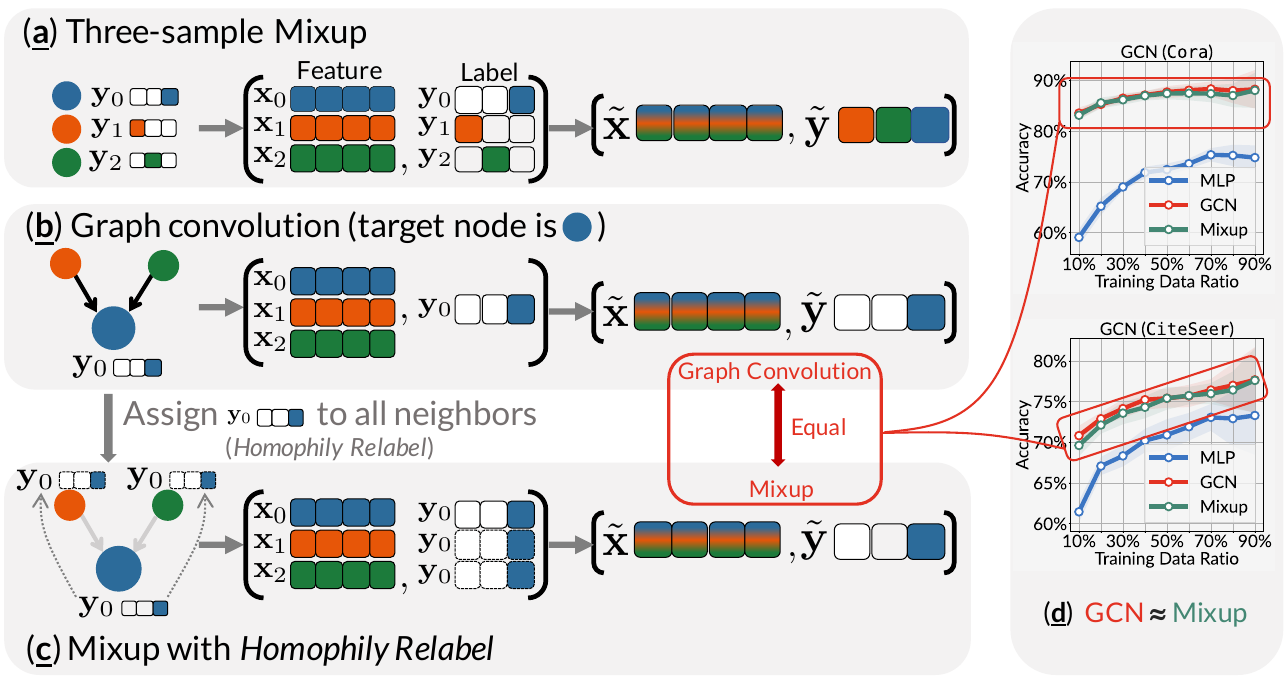}\vspace{-5pt}
    \caption{Graph convolution is Mixup. \textbf{(\underline{a})} illustrates the basic idea of Mixup: averaging the features and one-hot labels of multiple samples  ($\bluenode$, $\greennode, \orangenode$). \textbf{(\underline{b})} shows the graph convolution operation where the feature of the target node ($\bluenode$) is the weighted average of the features of all its neighbors. \textbf{(\underline{b}) $\rightarrow$ (\underline{c})} shows that graph convolution is Mixup if we assign the label ($\bluelabel$) of the target node ($\bluenode$) to all of its neighbors ($\greennode, \orangenode$). \textbf{(\underline{d})}  shows that Mixup is empirically equivalent to GCN.
    }\label{fig:intro} 
\end{figure}

In this paper, we answer this question by establishing the connection between graph convolutions and Mixup, and further understanding the graph neural networks through the lens of Mixup. We show that graph convolutions are intrinsically equivalent to Mixup by rewriting \cref{eq:gcn} as follows:
\begin{equation}\label{eq:gcn_2}
\begin{split}
    (\Tilde{\mathbf{x}}, \Tilde{\mathbf{y}}) = \left(\frac{1}{|\mathcal{N}_i|} \sum_{k \in \mathcal{N}_i} \mathbf{x}_k, {\color{red}\mathbf{y}_i}\right) = \left( \sum_{k \in \mathcal{N}_i} \frac{1}{|\mathcal{N}_i|} \mathbf{x}_k, \sum_{k \in \mathcal{N}_i} \frac{1}{|\mathcal{N}_i|} {\color{red} \mathbf{y}_i}\right) \overset{\lambda_i=\frac{1}{|\mathcal{N}_i|}}{=} \left( \sum_{k \in \mathcal{N}_i} \lambda_i \mathbf{x}_k, \sum_{k \in \mathcal{N}_i} \lambda_i {\color{red} \mathbf{y}_i}\right),\nonumber
\end{split}
\end{equation}
where $\mathbf{x}_i$ and ${\color{red} \mathbf{y}_i}$ are the feature and label of the target node $n_i$. This equation states that graph convolution is equivalent to Mixup if we assign the ${\color{red} \mathbf{y}_i}$ to all the neighbors of node $n_i$ in set $\mathcal{N}_i$ 

To demonstrate the equivalence between graph convolutions and Mixup, we begin by illustrating that a one-layer graph convolutional network (GCN)~\citep{kipf2016variational} can be transformed into an input Mixup. A two-layer GCN can be expressed as a hybrid of input and manifold Mixup~\citep{verma2019manifold}. Similarly, simplifing graph convolution (SGC)~\citep{wu2019simplifying} can be reformulated as an input Mixup. We thus establish the mathematical equivalence between graph convolutions and Mixup, under two mild and reasonable modifications: 1) assign the target node's label to neighbors in the training time (referred to as \textit{Homophily Relabel}); 2) perform feature mixup in the test time (referred to as \textit{Test-Time Mixup}).

We further investigate the modifications required for the equivalence between graph convolution and Mixup, focusing on the effect of \textit{Homophily Relabel} and \textit{Test-Time Mixup}. To explore \textit{Homophily Relabel}, we demonstrate that training an MLP with \textit{Homophily Relabel} (called \hmlp) is equivalent to training GCN in terms of prediction accuracy. This finding provides a novel perspective for understanding the relationship between graph convolution and Mixup and suggests a new direction for designing efficient GNNs. To investigate Test-Time Mixup, we train GNNs without connection information, perform neighbor aggregation during inference, and find that this approach can achieve performance comparable to traditional GNNs. This result reveals that Test-Time Mixup can be a powerful alternative to traditional GNNs in some scenarios, and suggests that Mixup may have a broader range of applications beyond traditional data augmentation. Our investigation on \textit{Homophily Relabeling} and \textit{Test-Time Mixup} provides valuable insights into the theoretical properties of GNNs and their potential applications in practice. We highlight our \textbf{contributions} as follows:

\begin{itemize}[leftmargin=0.8cm, itemindent=.0cm, itemsep=0.0cm, topsep=0.0cm]
    \item We establish for the first time connection between graph convolution and Mixup, showing that graph convolutions are mathematically and empirically equivalent to Mixup. This simple yet novel finding potentially opens the door toward a deeper understanding of GNNs.
    \item Given that both graph convolution and Mixup perform a weighted average on the features of samples, we reveal that graph convolution is conditionally equivalent to Mixup by assigning the label of a target node to all its neighbors during training time (\textit{Homophily Relabel}). Also, we also reveal that in the test time, graph convolution is also Mixup (\textit{Test-Time Mixup}).
    \item Based on \textit{Homophily Relabel} and \textit{Test-Time Mixup}, we propose two variants of MLPs based on the Mixup strategy, namely \hmlp and \tmlp, that can match the performance of GNNs.
\end{itemize}

\textbf{Related Work.} Graph neural networks are widely adopted in various graph applications, including social network analysis~\citep{fan2019graph}, recommendation~\citep{wei2022contrastive,deng2022graph,cai2023lightgcl,tang2022friend}, knowledge graph~\citep{cao2023autotransfer,zhao2023learning}, molecular analysis~\citep{sun2022does,zhu2023mathcalognn,zhang2023molecule,wang2023retrievalbased,corso2023diffdock,liao2023equiformer,xia2023molebert,hladi2023matching}, drug discovery~\citep{sun2022does,zhang2023ebind}, link prediction~\citep{chamberlain2023graph,cong2023do} and others~\citep{chen2023on}. Understanding the generalization and working mechanism of graph neural networks is still in its infancy~\citep{garg2020generalization,zhang2023rethinking,yang2022graph,baranwal2023effects}. The previous work attempts to understand graph neural networks from different perspectives, such as signal processing~\citep{nt2019revisiting,bo2021beyond,bianchi2021graph}, gradient flow~\citep{di2022graph}, dynamic programming\citep{dudzik2022graph}, neural tangent kernels~\citep{yang2022graph,du2019graph,sabanayagam2022representation} and influence function~\citep{chen2023characterizing}. There is also a line of works that analyzes the connection between GNNs and MLPs~\citep{baranwal2023effects,han2022mlpinit,yang2022graph,tian2023learning}, which is similar to the proposed method in~\cref{sec:homo}. In this work, we understand graph neural networks through a fresh perspective, Mixup. We believe that this work will inspire further research and lead to the development of new techniques to improve the performance and interpretability of GNNs. In parallel, Mixup~\citep{zhang2018mixup} and its variants~\citep{verma2019manifold,yun2019cutmix,kim2020puzzle,kim2021co} have emerged as a popular data augmentation technique that improves the generalization performance of deep neural networks. More specifically, deep neural networks trained with Mixup achieve better generalization~\citep{chun2020empirical,zhang2020does,chidambaram2022towards}, calibration~\citep{thulasidasan2019mixup,zhang2022and}, adversarial robustness~\citep{pang2019mixup,archambault2019mixup,zhang2020does,lamb2019interpolated}, and explain ensemble~\citep{lopezpaz2023an}. More related work in~\cref{sec:app:rela}.

\textbf{The Scope of Our Work.}~\footnote{Note that our work only focuses on the node-wise predictions made by GCNs.} This work not only provides a fresh perspective for comprehending Graph Convolution through Mixup, but also makes valuable contributions to the practical and theoretical aspects of graph neural networks, facilitating efficient training and inference for GNNs when dealing with large-scale graph data. \underline{Practical Potentials}: Both \hmlp and \tmlp hold potential value for practical usage. \tmlp is training-efficient, as its backbone and training process are both MLP-based; however, the inference requires \textit{Test-Time Mixup} (neighbor aggregation), which can be time-consuming. This limitation also exists in previous works~\citep{han2022mlpinit,yang2022graph}. \hmlp, on the other hand, is both training and test efficient, as both processes are based on MLP. This suggests the potential of \hmlp for practical usage on large-scale graphs.
\underline{Theoretical Potentials:} \hmlp, derived from the Mixup, have the theoretical potential to understand the expressiveness of GNNs from the Mixup perspective. With the connection between Mixup and graph convolution, \hmlp goes beyond traditional methods to understand the learning capability of GNNs.

\section{Graph Convolution is Mixup}\label{sec:gnnmixup}
In this section, we reveal that the graph convolution is essentially equivalent to Mixup. We first present the notation used in this paper. Then we present the original graph convolution network (GCN)~\citep{kingma2015adam} and simplifying graph convolutional networks (SGC)~\citep{wu2019simplifying} can be expressed mathematically as a form of Mixup. Last, we present the main claim, i.e.,  graph convolution can be viewed as a special form of Mixup under mild and reasonable modifications.

\textbf{Notations.} We denote a graph as $\mathcal{G}(\mathcal{V}, \mathcal{E})$, where $\mathcal{V}$ is the node set and $\mathcal{E}$ is the edge set. The number of nodes is $N=|\mathcal{V}|$ and the number of edges is $M=|\mathcal{E}|$. We denote the node feature matrix as $\mathbf{X} = \{\mathbf{x}_1, \mathbf{x}_2,\cdots, \mathbf{x}_N\} \in \mathbb{R}^{N \times d}$, where $\mathbf{x}_i$ is the node feature for node $n_i$ and $d$ is the dimension of features. We denote the binary adjacency matrix as $\mathbf{A} \in \{0, 1\}^{N\times N}$, where $\mathbf{A}_{ij}=1$ if there is an edge between node $n_i$ and node $n_j$ in edge set $\mathcal{E}$, and $\mathbf{A}_{ij}=0$ otherwise. We denote the neighbor set of node $n_i$ as $\mathcal{N}_i$, and its $2$-hop neighbor set as $\mathcal{N}_i^2$. For the node classification task, we denote the prediction targets of nodes as one-hot labels $\mathbf{Y} \in \{0, 1\}^{N\times (C-1)}$, where $C$ is the number of classes. For graph convolution network, we use $\Tilde{\mathbf{A}}=\mathbf{D}^{-1}\mathbf{A}$ as the normalized adjacency matrix\footnote{Our following analysis can be easily generalized to other normalization methods, such as $\mathbf{D}^{-\frac12}\mathbf{A}\mathbf{D}^{-\frac12}$.}, where $\mathbf{D} \in \mathbb{R}^{N\times N}$ is diagonal degree matrix, and $\mathbf{D}_{ii}$ is the degree of node $n_i$. We use $\Tilde{\mathbf{a}}_i \in \mathbb{R}^{1\times N}$, $i$-th row of $\Tilde{\mathbf{A}}$, as the normalized adjacency vector of node $n_i$.

\subsection{Preliminaries}\label{sec:gnnmixup:mixup}
\textbf{Graph Convolution.} Graph Convolution Network (GCN)~\citep{kipf2016semi}, as the pioneering work of GNNs, proposes  $\hat{\mathbf{Y}} = \text{softmax}( \Tilde{\mathbf{A}} \cdot \sigma( \Tilde{\mathbf{A}} \cdot \mathbf{X} \cdot \mathbf{W}_1 ) \cdot \mathbf{W}_2)$, where $\mathbf{W}_1$ and $\mathbf{W}_2$ are the trainable weights of the layer one and two, respectively. Simplifying Graph Convolutional Networks (SGC)~\citep{wu2019simplifying} is proposed as $\hat{\mathbf{Y}} = \text{softmax}( \Tilde{\mathbf{A}} \cdot \Tilde{\mathbf{A}} \cdot \mathbf{X} \cdot \mathbf{W} ) $. In this work, we take these two widely used GNNs to show that graph convolution is essentially Mixup.

\textbf{Mixup.} The Mixup technique, introduced by ~\citet{zhang2018mixup}, is a simple yet effective data augmentation method to improve the generalization of deep learning models. The basic idea behind Mixup is to blend the features and one-hot labels of a random pair of samples to generate synthetic samples. The mathematical expression of the two-sample Mixup is as follows:
\begin{equation}\label{eq:gcnmlp}
\begin{split}
(\Tilde{\mathbf{x}}, \Tilde{\mathbf{y}}) = (\lambda \mathbf{x}_i + (1 - \lambda) \mathbf{x}_j,  \lambda \mathbf{y}_i + (1 - \lambda)\mathbf{y}_j) = (\lambda_1 \mathbf{x}_i + \lambda_2 \mathbf{x}_j,  \lambda_1 \mathbf{y}_i +\lambda_2 \mathbf{y}_j),
\end{split}
\end{equation}

where $\lambda_i + \lambda_j = 1$. Based on the two-sample Mixup, the multiple-sample Mixup is presented in \cref{eq:mixup}.
The mathematical expression presented above demonstrates that Mixup computes a weighted average of the features from multiple original samples to generate synthetic samples.

\subsection{Connecting Graph Convolution and Mixup}\label{sec:gnnmixup:gnn}
We demonstrate that graph convolution, using GCN and SGC as examples, is conditionally Mixup. To do this, we mathematically reformulate the expressions of GCN and SGC to a Mixup form, thereby illustrating that graph convolutions are indeed Mixup.

\textbf{One-layer GCN is Mixup} We begin our analysis by examining a simple graph convolution neural network~\citep{kipf2016semi} with one layer, referred to as the one-layer GCN. We demonstrate that the one-layer GCN can be mathematically understood as an implementation of the input Mixup technique. The expression of the one-layer GCN is given by: $\hat{\mathbf{Y}} = \text{softmax}( \Tilde{\mathbf{A}} \cdot \mathbf{X} \cdot \mathbf{W} )$ where $\mathbf{W}\in \mathbb{R}^{D\times C}$ is the trainable weight matrix. Let us focus on a single node $n_i$ in the graph. The predicted one-hot label for this node is given by $\hat{\mathbf{y}_i} = \text{softmax}( \Tilde{\mathbf{a}}_i \cdot \mathbf{X} \cdot \mathbf{W} )$,
where $\Tilde{\mathbf{a}}_i \in \mathbb{R}^{1\times N} $, $i$-th row of $\Tilde{\mathbf{A}}$, is the normalized adjacency vector of node $n_i$. $\mathcal{N}_i$ is the neighbors set of node $n_i$. We make the following observations:
\begin{itemize}[leftmargin=0.6cm, itemindent=.0cm, itemsep=0.0cm, topsep=0.0cm]
    \item $\Tilde{\mathbf{a}}_i \cdot \mathbf{X}$ results in a weighted sum of the features of the neighbors of node $n_i$, which is the multiple-sample Mixup. Explicitly, we can rewrite $\Tilde{\mathbf{a}}_i \cdot \mathbf{X}$ to $\Tilde{\mathbf{x}} = \frac{1}{|\mathcal{N}i|} \sum_{k \in \mathcal{N}_i} \mathbf{x}_k$.
    \item If we assign node $n_i$'s label to all its neighbors, the label $\mathbf{y}_i$ of node $n_i$ can be interpreted as a weighted average of the labels of its neighboring nodes, which is equivalent to performing Mixup on the labels.
    Thus we have $\Tilde{\mathbf{y}} = \mathbf{y}_i =  \frac{1}{|\mathcal{N}_i|} \sum_{k \in \mathcal{N}_i} \mathbf{y}_k$.
\end{itemize}

From the previous observations, we can see that one-layer GCN synthesizes a new Mixup sample for node $n_i$ by mixing up its neighbors' features and one-hot labels, which is defined as follows:
\begin{equation}
    (\Tilde{\mathbf{x}}, \Tilde{\mathbf{y}}) = \left(\frac{1}{|\mathcal{N}_i|} \sum_{k \in \mathcal{N}_i} \mathbf{x}_k, \frac{1}{|\mathcal{N}_i|} \sum_{k \in \mathcal{N}_i} \mathbf{y}_k\right).
\end{equation}
Therefore, we conclude that \textit{a one-layer GCN is a Mixup machine, which essentially synthesizes a new sample by averaging its neighbors' features and one-hot labels}.

\textbf{Two-layer GCN is Mixup} We extend our analysis to consider a two-layer GCN. The expression for two-layer GCN is given by $\hat{\mathbf{Y}} = \text{softmax}( \Tilde{\mathbf{A}} \cdot \sigma( \Tilde{\mathbf{A}} \cdot \mathbf{X} \cdot \mathbf{W}_1 ) \cdot \mathbf{W}_2)$, 
Let us focus on node $n_i$ in the graph. For the first layer of the two-layer GCN, we have $\mathbf{h}_i = \text{Relu}( \Tilde{\mathbf{a}}_i \cdot \mathbf{X} \cdot \mathbf{W}_1 ) \in \mathbb{R}^{1\times d}$,
where $\Tilde{\mathbf{a}}_i \in \mathbb{R}^{1\times N}$ is the normalized adjacency vector of node $n_i$. The first layer is the same as the one-layer GCN as discussed above, which can be regarded as a multiple-sample input Mixup.

For the second layer of the two-layer GCN, we have $\hat{\mathbf{y}}_i = \text{Softmax}( \Tilde{\mathbf{a}}_i \mathbf{H} \mathbf{W}_2 ) \in \mathbb{R}^{1\times C}$, $\mathbf{H} \in \mathbb{R}^{N\times d}$ is the hidden representation of all nodes obtained from the first layer, and $\mathbf{W}_2 \in \mathbb{R}^{d\times C}$ is the weight matrix. The predicted one-hot label for node $n_i$, $\hat{\mathbf{y}}_i$, is obtained through a softmax activation function. The $2$-nd layer can be regarded as multiple-sample manifold Mixup~\citep{verma2019manifold} in the following:
\begin{itemize}[leftmargin=0.8cm, itemindent=.0cm, itemsep=0.0cm, topsep=0.0cm]
    \item $\Tilde{\mathbf{a}}_i \cdot \mathbf{H}$ is the multiple-sample Mixup of the hidden representation of the neighbors of node $n_i$. We rewrite the $\Tilde{\mathbf{a}}_i \cdot \mathbf{H}$ to $\Tilde{\mathbf{x}} = \Tilde{\mathbf{a}}_i \cdot \mathbf{H} =  \frac{1}{|\mathcal{N}_i|} \sum_{k \in \mathcal{N}_i} \mathbf{h}_k$.
\end{itemize}

Therefore, we conclude that \textit{a multi-layer GCN is a hybrid of input Mixup (first layer) and manifold Mixup (second layer)}.

\textbf{SGC is Mixup} The $2$-layer SGC architecture is represented by the following expression, $\hat{\mathbf{Y}} = \text{softmax}( \Tilde{\mathbf{A}} \cdot \Tilde{\mathbf{A}} \cdot \mathbf{X} \cdot \mathbf{W} )$. 
Similar to one-layer GCN, for the node $n_i$ in a 2-layer SGC, we have
\begin{equation}\label{equ:gcn6}
\begin{split}
\hat{\mathbf{y}_i} = \text{softmax}( \Tilde{\mathbf{a}}_i^2 \cdot \mathbf{X} \cdot \mathbf{W} ), \mathcal{L} = \text{Cross-Entropy}(\hat{\mathbf{y}_i}, \mathbf{y}_i),
\end{split}
\end{equation}
where $\Tilde{\mathbf{a}}_i^2 \in \mathbb{R}^{1\times N} $ is the adjacency vector within $2$-hop neighbors of node $n_i$. The $2$-hop neighbor set of node $n_i$ is represented by $\mathcal{N}^2_i$, 
Hereby we have
\begin{itemize}[leftmargin=0.8cm, itemindent=.0cm, itemsep=0.0cm, topsep=0.0cm]
\item $\Tilde{\mathbf{a}}^2_i \cdot \mathbf{X}$ is the multiple Mixup of the features of the neighbors of node $n_i$. We rewrite the $\Tilde{\mathbf{a}}i \cdot \mathbf{X}$ to $\Tilde{\mathbf{x}} = \Tilde{\mathbf{a}}^2_i \cdot \mathbf{X} = \frac{1}{|\mathcal{N}^2_i|} \sum{k \in \mathcal{N}^2_i} \mathbf{x}_k$.
\end{itemize}

From the above, we have a Mixup of samples for node $n_i$ is
\begin{equation}
(\Tilde{\mathbf{x}}, \Tilde{\mathbf{y}}) = \left(\frac{1}{|\mathcal{N}^2_i|} \sum_{k \in \mathcal{N}^2_i} \mathbf{x}_k, ~~\frac{1}{|\mathcal{N}^2_i|} \sum_{k \in \mathcal{N}^2_i} \mathbf{y}_k\right).
\end{equation}
Thus we conclude that \textit{an SGC is an input Mixup machine}.

\subsection{Graph Convolution is (\textit{Conditionally}) Mixup}\label{sec:gnnmixup:main}

It is straightforward to derive the Mixup form of $1$-layer, $2$-layer GCN, and SGC as discussed above. This leads to the conclusion that $1$-layer, $2$-layer GCN, and SGC can all be reformulated in the form of Mixup. This establishes a mathematical connection between graph neural networks and the Mixup method. Building on these findings, in this section, we introduce our main contribution as follows:

\begin{applebox}{Main Results}
  Graph convolution is Train- and Test-Time Mixup, under two mild and reasonable modifications: \textit{Homophily Relabel} and  \textit{Test-time Mixup}.
\end{applebox}

The two primary differences between GNNs and Mixup are as follows:

\begin{itemize}[leftmargin=0.6cm, itemindent=.0cm, itemsep=0.0cm, topsep=0.0cm]
    \item \textit{\textbf{Homophily Relabel}} At training time, if we assign the label of the target node to all its neighbors, the Graph Neural Network can be naturally rewritten as a form of Mixup.
    \item \textit{\textbf{Test-Time Mixup}} At test time, the GNNs perform feature mixing on the nodes and then use the mixed feature for the inference.
\end{itemize}

Both of the above modifications are mild and reasonable and have practical implications for GNN design and analysis. The \textit{Homophily Relabel} operation can be understood as imposing the same label on the target node and all its neighbors, which corresponds to the homophily assumption for graph neural networks. The homophily assumption posits that nodes with similar features should have the same label, which is a common assumption in many real-world graph applications.  On the other hand, \textit{Test-time Mixup} can be understood as Mixup at the test time by mixing the features from neighbors for prediction. We examine the above differences in depth in \cref{sec:homo} and \cref{sec:testmixup}.

\subsection{Discussion}\label{sec:dis}

\textbf{Comparison to Other Work.} Hereby, we compare our work and previous work. \citet{yang2022graph,han2022mlpinit} propose to train an MLP and transfer the converged weight from MLP to GNN, which can achieve comparable performance to GCN. In this work, these two methods are all included in part of our work, \textit{Test-Time Mixup}. Different from previous work, we provide a fresh understanding of this phenomena by connecting graph convolution to Mixup and also derive a \tmlp~\cref{sec:app:exp:tmlp} to implement \textit{Test-Time Mixup}. \citet{baranwal2022effects} understand the effect of graph convolution by providing a rigorous theoretical analysis based on a synthetic graph. Different from this work,  our work understands graph convolution with a well-studied technique, Mixup.

\begin{wraptable}[5]{R}{0.45\textwidth}
    \vspace{-23pt}
    \centering
    \fontsize{9}{9}\selectfont
    \setlength{\tabcolsep}{3pt}
    \caption{The running time of one epoch comparison of original MLP vs. GNN.}\label{tab:mlp_gnn}
    \begin{tabular}{l|c|c}
    \toprule
                      & \textbf{MLP} & \textbf{GNN} \\ \midrule
    \textbf{Running time} & 1.8396±0.0264 & 174.1459±7.0611 \\ 
    \bottomrule
    \end{tabular}

\end{wraptable}

\textbf{Why MLP is more efficient to GNN:} GNN is $\hat{\mathbf{Y}} = \text{softmax}( \Tilde{\mathbf{A}} \cdot \mathbf{X} \cdot \mathbf{W} )$, where $\mathbf{A}$ is a large sparse matrix, resulting in significant computation time. This is well studied in previous work \cite{zhang2021graph,yang2022graph}.  We also conducted experiments to measure the actual runtime of MLP and GNN. The results demonstrate the efficiency of MLP compared to GNN. With mixup, the number of data points will double. The running time of MLP will at most double, resulting in a running time ratio of MLP to GNN = (1.8396 * 2) : 174.1459 = 1 : 47.

\section{Is \textit{Homophily Relabel} Equivalent to GCNs Training ?}\label{sec:homo}
In this section, we conduct experiments to examine the \textit{Homophily Relabel} proposed in \cref{sec:gnnmixup:main}. The empirical evidence substantiates our claims, emphasizing the significant effect of Homophily Relabel on GNNs. This understanding can facilitate the design of an MLP equivalent that competes with the performance of GNNs. Note that we explore the transductive setting in our experiments.

\subsection{\textit{Homophily Relabel} Yields a Training-Equivalent MLP to GCN}
\begin{wrapfigure}[10]{R}{0.25\textwidth}
    \centering
    \vspace{-15pt}
    \includegraphics[width=0.2\textwidth]{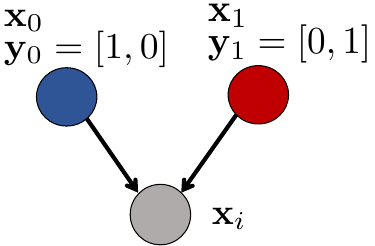}\vspace{-5pt}
    \caption{The example graphs. $\mathbf{x}_i$ is the target node, The loss of $\mathbf{x}_i$ that connected two nodes with different labels.}\label{fig:3nodes}
\end{wrapfigure}
We utilize the simple example graph to demonstrate the calculation of the loss function in a GCN in the transductive setting. The example graph, in \cref{fig:3nodes}, comprises three nodes. The {\color{myblue}blue} and {\color{myred}red} nodes belong to the training set, while the gray node belongs to the test set. In the transductive scenario, the loss function is computed for the entire graph during the training phase, incorporating both the training and test nodes. 
In our simple example graph, the loss function will be calculated on the blue and red nodes with their actual labels.  Notably, the prediction for the gray node, despite it being part of the test set, will also contribute to the overall loss calculation in the transductive setting. For node $n_i$, the prediction $ (p_0, p_1)$, the loss of the example graph will be as follows
\begin{equation}
\begin{split}
    L_{\mathcal{G}} &= \text{cross-entropy}(\mathbf{Y}, \hat{\mathbf{Y}}) =\text{cross-entropy}(\mathbf{y}_0, \hat{\mathbf{y}_0}) + \text{cross-entropy}(\mathbf{y}_1, \hat{\mathbf{y}_1})\\
    &= { \color{myred} -y_{00} \log (p_0) - y_{01} \log (p_1) }  {\color{myblue} -y_{10} \log (p_0) - y_{11} \log (p_1) }\\
    &= { \color{myred} - 1 * \log (p_0) - 0 * \log (p_1) }  {\color{myblue} - 0 * \log (p_0) - 1 * \log (p_1) }\\
    &= { \color{myred} - 1 * \log (p_0) }  {\color{myblue} - 1 * \log (p_1) } = -2*({ \color{myred} 0.5 * \log (p_0) }  {\color{myblue} + 0.5 * \log (p_1) } ).
\end{split}
\end{equation}

where $y_{00}$/$y_{01}$ is the 0th/1st elements in the one-hot label $\mathbf{y}_0$ for node $n_0$. The above analysis shows that the actual training label for target node $n_i$ is $(0.5, 0.5)$. From the Mixup perspective, in an ideally well-trained model utilizing a two-sample Mixup, as shown in previous work~\citep{guo2019mixup,pang2019mixup}, the following approximation will hold $f(\lambda_k x + (1-\lambda_k)x_k) \approx \lambda_kf(x)+(1-\lambda_k)f(x_k)$. Intuitively, this approximation holds because the model is trained with mixup examples $(x_{new}, y_{new}) = (\lambda x_i + (1 - \lambda) x_j, \lambda y_i + (1 - \lambda) y_j)$. The linear correlation between $x_{new}$ and $y_{new}$ leads to a model that can be approximated by a linear function.
If we regard the neighbor aggregation as Mixup, we can easily derive that
\begin{equation}\label{equ:mix0}
  \begin{split}
       \hat{\mathbf{y}}_0 = f\left( \frac{\mathbf{x}_0 + \mathbf{x}_i}{2} \right) =\frac{1}{2} \left( f( \mathbf{x}_0) + f(\mathbf{x}_i) \right), \hat{\mathbf{y}}_1 = f\left( \frac{\mathbf{x}_1 + \mathbf{x}_i}{2} \right) =\frac{1}{2} \left( f( \mathbf{x}_1) + f(\mathbf{x}_i) \right).
  \end{split}
\end{equation}
From the above two equations, we can see that $f(x_i)$ are trained with the labels of all its neighbors $\mathbf{x}_0$ and $\mathbf{x}_1$. Thus in the next, \textit{we propose explicitly training the nodes in the test set with the label of their neighbors if they are in the training set.} In our example graph, we explicitly train the gray node with label $(0.5, 0.5)$

\paragraph{\hmlp} Based on the above analysis, we proposed Homophily Relabel MLP (\hmlp), which achieves the comparable performance of GCN via training an MLP as a backbone with \textit{Homophily Relabel}. In detail, the proposed method \hmlp has two major steps: 1) relabel all the nodes in the graph with the mixed label $\hat{\mathbf{Y}} = \mathbf{A} \cdot \mathbf{Y}$. 2) train an MLP on all the nodes with $\hat{\mathbf{Y}}$ on relabeled graph, only using node features. We illustrate our proposed \hmlp with the example graph in \cref{fig:3nodes}. 1) we relabel all the nodes using \textit{Homophily Relabel}, the node with new label will be $(\mathbf{x}_0, (1,0)), (\mathbf{x}_1, (0,1)), (\mathbf{x}_i, (0.5,0.5))$  2) then train the \hmlp with these data samples. The test sample will be $(\mathbf{x}_i, ?)$ during the test time.

\subsection{Can \hmlp Achieve Comparable Performance to GCN?}
\begin{wraptable}[9]{R}{0.45\textwidth}
    \vspace{-18pt}
    \centering
    \fontsize{9}{9}\selectfont
    \setlength{\tabcolsep}{3pt}
    \caption{Performance comparison of GCN and \hmlp.}\label{tab:graphconv_mixup_hmlp}
      \begin{tabular}{lc|cccc}
      \toprule
      Datasets          &MLP                 &GCN                 &\hmlp                 \\ \midrule
      \cora             &\ms{73.57}{0.98}    &\ms{88.04}{1.26}    &\ms{86.42}{1.78}      \\
      \citeseer         &\ms{71.90}{1.69}    &\ms{75.69}{1.36}    &\ms{75.94}{1.54}      \\ 
      \pubmed           &\ms{86.90}{0.74}    &\ms{87.94}{0.64}    &\ms{88.34}{0.48}      \\ \midrule\rowcolor{gray!20}
      Average           &$77.45$             &$\textbf{83.89}$             &$\textbf{83.57}$           \\ 
      \bottomrule
      \end{tabular}
\end{wraptable}

\begin{wrapfigure}[11]{R}{0.45\textwidth}
    \vspace{-5pt}
    \includegraphics[width=0.999\linewidth]{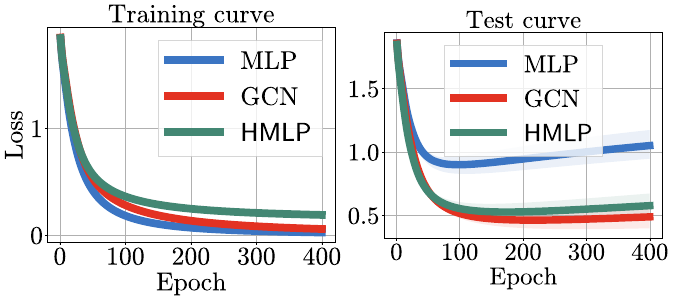}\vspace{-5pt}
    \caption{The training and test curves of GCN and \hmlp.}\label{fig:hmlp_curve}
\end{wrapfigure}

\begin{figure}[!t]
  \centering
  \includegraphics[width=0.9999\textwidth]{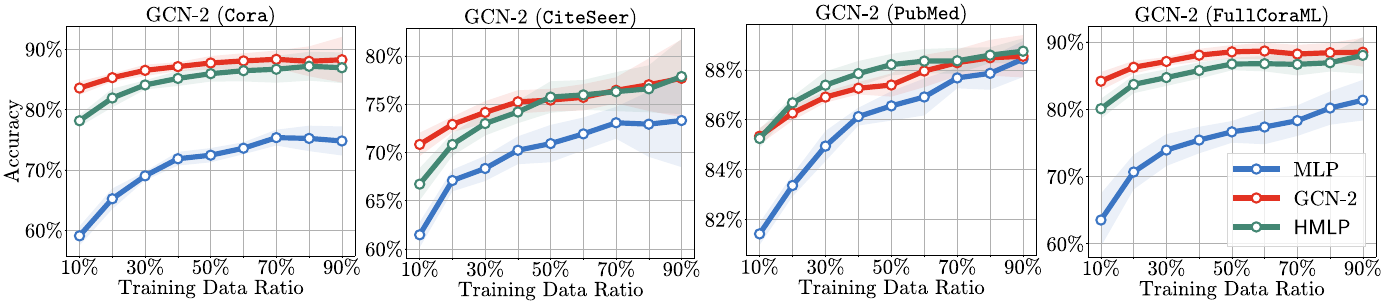}\vspace{-5pt}
  \caption{The performance comparison of the {\color{myred}GCN}, {\color{myblue}MLP} and {\color{mygreen}\hmlp} (Ours). The x-axis represents the ratio of training data, and the y-axis represents the accuracy of classification. The results show that our proposed method ({\color{mygreen}\hmlp}) achieves comparable performance to {\color{myred}GCN}. Note that the architecture of our method in train and test time are both MLP. More experimental results on other datasets and GNNs are presented in \cref{sec:app:exp:hmlp}.}\label{fig:pref_hmlp}
\end{figure}

To verify the proposed \hmlp can achieve comparable performance to the original GCN, we train the one-layer GCN (GCN-1), two-layer GCN (GCN-2), SGC, and \hmlp on the training set and report the test accuracy based on the highest accuracy achieved on the validation set. We experimented with different data splits of train/validation/test (the training data ratio span from $10\%-90\%$), and we also conducted experiments with the public split on \cora, \citeseer, and \pubmed datasets. We present the results in \cref{tab:graphconv_mixup_hmlp} and \cref{fig:pref_hmlp}. We also present the training and test curves of the MLP. GCN and \hmlp in \cref{fig:hmlp_curve}.

From the results in \cref{tab:graphconv_mixup_hmlp} and \cref{fig:pref_hmlp}, \ding{202} \textbf{\hmlp achieves comparable performance to GCN when the ratio of training to test data is large.} For the \cref{fig:pref_hmlp}, the performance of \hmlp is significantly better than MLP, and on par with GCN, especially when the training data ratio is large. The results also show that when the training data ratio is small, \hmlp performs worse than GCN. The training and test curves show that \hmlp achieve the similar test curves ({\color{myred}red} and {\color{mygreen}green} lines in the right subfigure) even though the training curve are not the same. This may be because the labels of nodes are changed.

\section{How Does \textit{Test-time Mixup} Affect GCN Inference?}\label{sec:testmixup}

In this section, we aim to investigate how \textit{Test-time Mixup} affects the GCN inference process. We conduct an empirical study to understand the functionality of \textit{Test-time Mixup} (neighbor aggregation during the test time).  To this end, we propose an MLP involving \textit{Test-time Mixup} only (\tmlp). 

\textbf{\tmlp} The proposed method \tmlp follows the steps: 1) we train an MLP using only the node features of the training set. 2) we employ the well-trained MLP for inference with neighbor aggregation during testing. This paradigm has also been investigated in previous works~\citep{han2022mlpinit,yang2022graph}. We call this method \tmlp.  We illustrate the \tmlp with the example graph in \cref{fig:3nodes}. Specifically, we train the model using the following training examples, $(\mathbf{x}_0, (1,0)), (\mathbf{x}_1, (0,1))$. Note that the backbone of \tmlp is an MLP. In the following, we explore the performance of \tmlp.

\subsection{How Does \tmlp Perform Compared to GCN?}

\begin{figure}[!t]
  \centering
  \includegraphics[width=0.9999\textwidth]{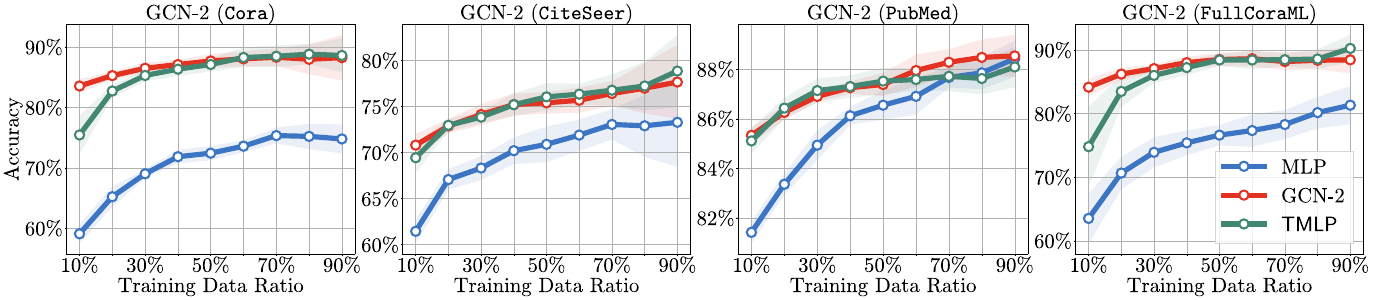}\vspace{-5pt}
  \caption{The performance comparison of the {\color{myred}GCNs}, {\color{myblue}MLP} and {\color{mygreen}\tmlp} (Ours). GCN-1 is a one-layer GCN, and GCN-2 is a two-layer GCN. The results show that \tmlp achieves a comparable performance to GCN and SGC. Note that the architecture is essentially MLP in training time and GCN in the test time. More results on other datasets and GNNs are presented in \cref{sec:app:exp:tmlp}.}\label{fig:pref_tmlp}
\end{figure}

In the section, we conducted experiments to verify the effect of \textit{Test-Time Mixup}. In the experiments, during the training, we only use the node feature to train an MLP,  and then during the test time, we use the well-trained MLP with the neighbor aggregation to perform the inference. We present the results of varying training data ratio in \cref{fig:pref_tmlp}, and we also present the accuracy in ~\cref{tab:graphconv_mixup_tmlp} when the training data ratio is $0.6$.  Additionally, we compare the learned representation of GNNs and \tmlp using t-SNE~\cite{van2008visualizing}. From these results, we make the following observations:

\begin{wraptable}[9]{R}{0.45\textwidth}
    \vspace{-15pt}
    \centering
    \fontsize{9}{9}\selectfont
    \setlength{\tabcolsep}{3pt}
    \caption{Performance comparison of GCN and \hmlp.}\label{tab:graphconv_mixup_tmlp}
      \begin{tabular}{lc|cccc}
      \toprule
      Datasets          &MLP                 &GCN                 &\tmlp                 \\ \midrule
      \cora             &\ms{73.57}{0.98}    &\ms{88.04}{1.26}    &\ms{88.26}{1.60}      \\
      \citeseer         &\ms{71.90}{1.69}    &\ms{75.69}{1.36}    &\ms{76.35}{1.13}       \\ 
      \pubmed           &\ms{86.90}{0.74}    &\ms{87.94}{0.64}    &\ms{87.58}{0.44}      \\ \midrule\rowcolor{gray!20}
      Average           &$77.45$             &$\textbf{83.89}$             &$\textbf{84.06}$           \\ 
      \bottomrule
      \end{tabular}
\end{wraptable}

From these results, we make the following observations: \ding{203} \textbf{\tmlp almost achieves comparable performance to GNNs.} In the \cref{fig:pref_tmlp}, the performance of \tmlp is significantly better than MLP and on par with GCN, especially when the training data ratio is large. In some datasets (i.e., \cora, \coraml), the performance of \tmlp is slightly worse than GCN but still significantly better than MLP. It is worth noting that our proposed \tmlp is only trained with the node features, ignoring the connection information. \cref{tab:graphconv_mixup_tmlp} show the average accuracy ($84.06$) of \tmlp is comparable to that of GCN ($83.89$).

\begin{wrapfigure}[10]{R}{0.450\textwidth}
    \vspace{-10pt}
    \centering
    \includegraphics[width=0.95\linewidth]{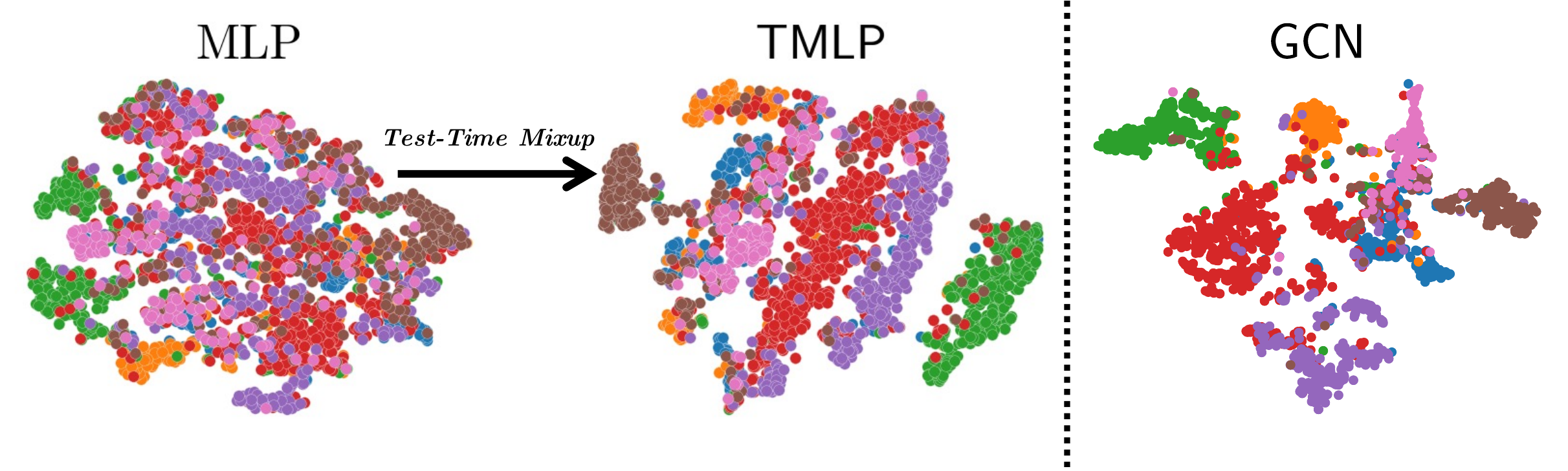}\vspace{-5pt}
    \caption{Visualization of node representations learned by MLP and \tmlp. Node representations from the same class, after \textit{Test-Time Mixup}, become clustered together.
    }\label{fig:emb_change}
\end{wrapfigure}
Based on the node representation, \ding{204} \textbf{\textit{Test-Time Mixup} make the node representation more discriminative than MLP.}  The visualization clearly illustrates that the node representations within the same class, as learned by \tmlp, are clustered together. This clustering pattern demonstrates the advanced predictive power of \tmlp, as it appears to have a keen ability to distinguish similar group classes effectively. Consequently, this capacity allows \tmlp to deliver superior classification performance. The node representation clearly shows what \tmlp does from the representation perspective.

\subsection{How Does \textit{Test-time Mixup} Benefit GCN? Investigating Decision Boundary}\label{sec:testmixup:boundary}

In this section, we evaluate the effect of \textit{Test-Time Mixup} on the decision boundary change from MLP to \tmlp. To do so, we compare the distance between node features and the decision boundary of MLP before and after applying \textit{Test-Time Mixup}. After training a model with training-time data mixup, we evaluate its decision boundary, which separates different classes in the feature space. We then apply test-time mixup techniques to node features and reassess the decision boundary. Our findings indicate that \ding{205} \textbf{\textit{Test-Time Mixup} can cause node features to move away from the learned boundary}, enhancing model robustness to unseen data. However, it's crucial to strike a balance between \textit{Test-Time Mixup} and model performance, as overdoing Mixup may reduce accuracy.

\begin{figure}
    \centering
    \includegraphics[width=0.95\textwidth]{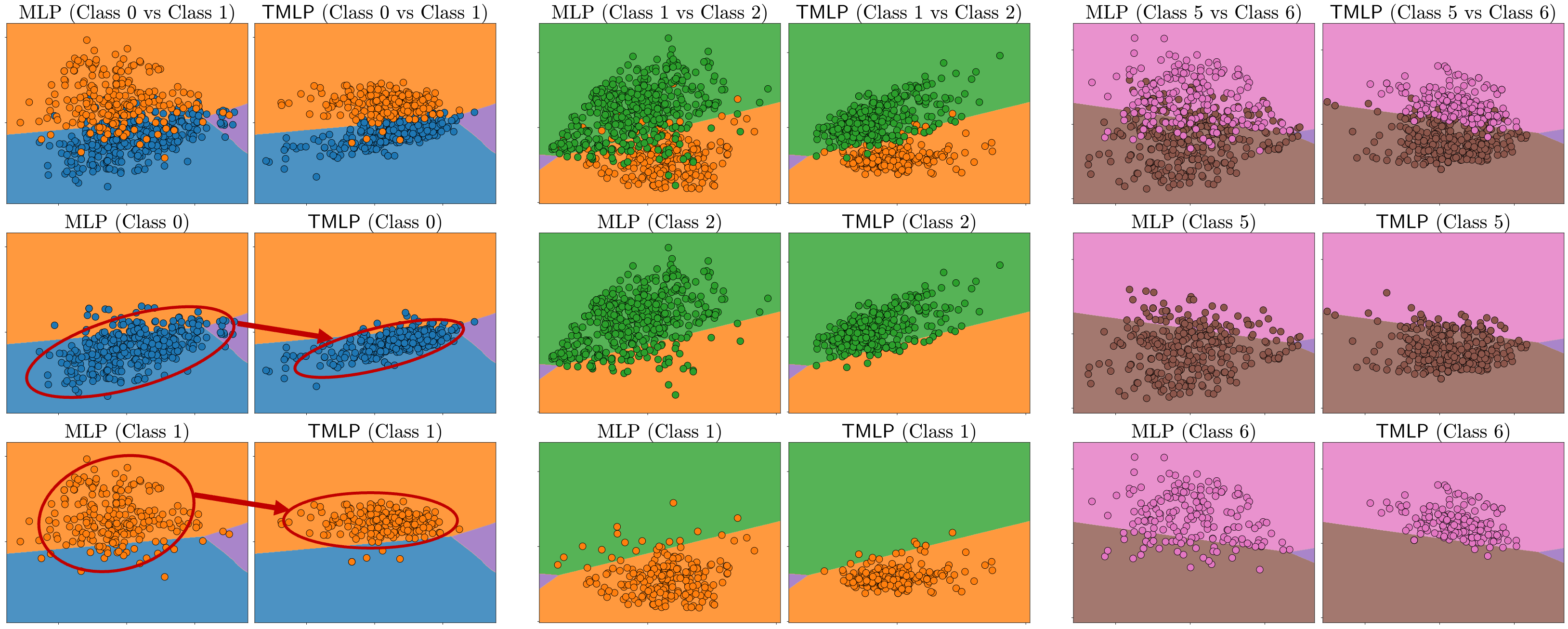}\vspace{-5pt}
    \caption{Decision Boundary of MLP and \tmlp. \tmlp can cause node features to move away from the learned boundary, as shown with $\dblegend$. More experimental details are in \cref{sec:app:exp:db}}\label{fig:decision_boundary}
\end{figure}

\section{Unifying \hmlp and \tmlp: Verifying the Main Results}

\begin{figure}
    \centering
    \includegraphics[width=0.999\textwidth]{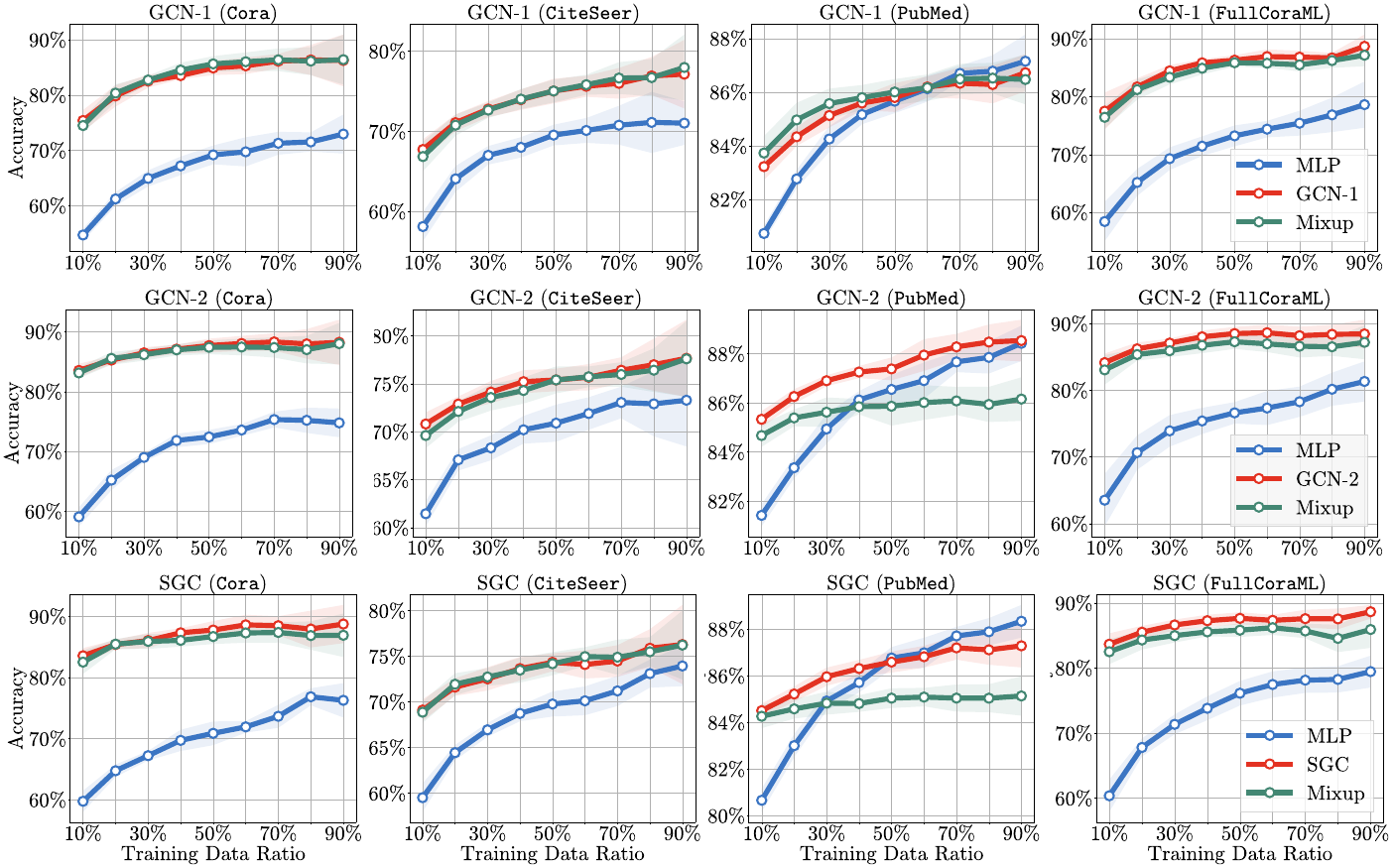}\vspace{-5pt}
    \caption{Unifying \hmlp and \tmlp. these methods together achieve performance comparable to GNNs (excluding the \pubmed dataset, as the MLPs also exhibit better performance than GNNs.). }\label{fig:mlp_all}
\end{figure}

In this section, we unify \hmlp and \tmlp into a single MLP, which integrates both \textit{Homophily Relabel} and \textit{Test-time Mixup}. They are supposed to be equivalent to graph convolution. To unify these two MLPs, we 1) relabel all the nodes in the graph with the mixed label $\hat{\mathbf{Y}} = \mathbf{A} \cdot \mathbf{Y}$. 2) train an MLP on all the nodes with $\hat{\mathbf{Y}}$ on relabeled graph, only using node features. 3)  employ the well-trained MLP for inference with neighbor aggregation during testing. To test the efficacy of this combined approach, we perform experiments comparing the performance of the unified MLP with that of individual \hmlp and \tmlp, as well as other GCN. The experimental setup remains consistent with the previous setting with \hmlp and \tmlp. We present the results in \cref{fig:mlp_all}.  \ding{206} \textbf{Unifying \hmlp and \tmlp can achieve comparable performance to original GNNs.}

\section{Discussion and Future Work}
In this paper, we reveal the relation between graph neural networks and Mixup that graph neural networks are essentially Mixup, in the sense that they can be seen as a continuous interpolation between nodes in the graph. This equivalence allows for a new interpretation of GNNs as a form of regularization, and provides a new perspective on the design of GNNs.

\begin{itemize}[leftmargin=0.8cm, itemindent=.0cm, itemsep=0.0cm, topsep=0.0cm]
    \item \textbf{Efficient Alternative Training and Inference Method for GNNs on Large Graph Data.} \hmlp presents an efficient solution for both training and testing on large-scale graphs, as it completely eliminates the need for connection information during both stages. \tmlp, though also MLP-based, emphasizes training efficiency. The improved training efficiency of \tmlp can still be advantageous for reducing the training cost on large graph.
    \item \textbf{Opening a New Door for Understanding Graph Convolution.} The introduction of mixup, a simple node-level operation, offers a novel perspective on graph convolution by representing it in a mixup form. Traditional graph convolution employs the adjacency matrix of the entire graph, a concept that can be complex to comprehend or articulate. By approaching the understanding of graph neural networks from the entire graph level, we can demystify these complexities. This presents an essential avenue for exploration in future work related to our paper.
\end{itemize}

\textbf{Future Work.} Based on the above practical and theoretical potentials of the proposed methods in this paper. Here are some potential future works:  1) improve \hmlp to make it more practical for large-scale graphs and the performance when the training data ratio is small.  2) The well-studied Mixup strategy would be helpful for understanding the expressiveness of graph neural networks.

\clearpage
\section*{Acknowledgements}
This work was done while Xiaotian Han was an intern at Meta. We are grateful for the detailed reviews from the reviewers. Their thoughtful comments and suggestions were invaluable in strengthening this paper. Portions of this research were conducted with high performance research computing resources provided by Texas A\&M University (https://hprc.tamu.edu). This work was supported in part by the National Science Foundation (NSF) IIS-2224843 and by the US Department of Transportation (USDOT) Tier\-1 University Transportation Center (UTC) Transportation Cybersecurity Center for Advanced Research and Education (CYBER-CARE) grant \#69A3552348332.

\bibliography{ref}

\begin{thebibliography}{79}
\providecommand{\natexlab}[1]{#1}
\providecommand{\url}[1]{\texttt{#1}}
\expandafter\ifx\csname urlstyle\endcsname\relax
  \providecommand{\doi}[1]{doi: #1}\else
  \providecommand{\doi}{doi: \begingroup \urlstyle{rm}\Url}\fi

\bibitem[Archambault et~al.(2019)Archambault, Mao, Guo, and Zhang]{archambault2019mixup}
Guillaume~P Archambault, Yongyi Mao, Hongyu Guo, and Richong Zhang.
\newblock Mixup as directional adversarial training.
\newblock \emph{arXiv preprint arXiv:1906.06875}, 2019.

\bibitem[Baranwal et~al.(2022)Baranwal, Fountoulakis, and Jagannath]{baranwal2022effects}
Aseem Baranwal, Kimon Fountoulakis, and Aukosh Jagannath.
\newblock Effects of graph convolutions in deep networks.
\newblock \emph{arXiv preprint arXiv:2204.09297}, 2022.

\bibitem[Baranwal et~al.(2023)Baranwal, Fountoulakis, and Jagannath]{baranwal2023effects}
Aseem Baranwal, Kimon Fountoulakis, and Aukosh Jagannath.
\newblock Effects of graph convolutions in multi-layer networks.
\newblock In \emph{International Conference on Learning Representations}, 2023.
\newblock URL \url{https://openreview.net/forum?id=P-73JPgRs0R}.

\bibitem[Bianchi et~al.(2021)Bianchi, Grattarola, Livi, and Alippi]{bianchi2021graph}
Filippo~Maria Bianchi, Daniele Grattarola, Lorenzo Livi, and Cesare Alippi.
\newblock Graph neural networks with convolutional arma filters.
\newblock \emph{IEEE transactions on pattern analysis and machine intelligence}, 44\penalty0 (7):\penalty0 3496--3507, 2021.

\bibitem[Bo et~al.(2021)Bo, Wang, Shi, and Shen]{bo2021beyond}
Deyu Bo, Xiao Wang, Chuan Shi, and Huawei Shen.
\newblock Beyond low-frequency information in graph convolutional networks.
\newblock In \emph{Proceedings of the AAAI Conference on Artificial Intelligence}, volume~35, pp.\  3950--3957, 2021.

\bibitem[Cai et~al.(2020)Cai, Wang, and Wang]{cai2020graph}
Chen Cai, Dingkang Wang, and Yusu Wang.
\newblock Graph coarsening with neural networks.
\newblock In \emph{International Conference on Learning Representations}, 2020.

\bibitem[Cai et~al.(2023)Cai, Huang, Xia, and Ren]{cai2023lightgcl}
Xuheng Cai, Chao Huang, Lianghao Xia, and Xubin Ren.
\newblock Light{GCL}: Simple yet effective graph contrastive learning for recommendation.
\newblock In \emph{The Eleventh International Conference on Learning Representations}, 2023.
\newblock URL \url{https://openreview.net/forum?id=FKXVK9dyMM}.

\bibitem[Cao et~al.(2023)Cao, You, Liu, and Leskovec]{cao2023autotransfer}
Kaidi Cao, Jiaxuan You, Jiaju Liu, and Jure Leskovec.
\newblock Autotransfer: Auto{ML} with knowledge transfer - an application to graph neural networks.
\newblock In \emph{International Conference on Learning Representations}, 2023.
\newblock URL \url{https://openreview.net/forum?id=y81ppNf_vg}.

\bibitem[Chamberlain et~al.(2023)Chamberlain, Shirobokov, Rossi, Frasca, Markovich, Hammerla, Bronstein, and Hansmire]{chamberlain2023graph}
Benjamin~Paul Chamberlain, Sergey Shirobokov, Emanuele Rossi, Fabrizio Frasca, Thomas Markovich, Nils~Yannick Hammerla, Michael~M. Bronstein, and Max Hansmire.
\newblock Graph neural networks for link prediction with subgraph sketching.
\newblock In \emph{International Conference on Learning Representations}, 2023.
\newblock URL \url{https://openreview.net/forum?id=m1oqEOAozQU}.

\bibitem[Chen et~al.(2020)Chen, Wei, Huang, Ding, and Li]{chen2020simple}
Ming Chen, Zhewei Wei, Zengfeng Huang, Bolin Ding, and Yaliang Li.
\newblock Simple and deep graph convolutional networks.
\newblock In \emph{International conference on machine learning}, pp.\  1725--1735. PMLR, 2020.

\bibitem[Chen et~al.(2023{\natexlab{a}})Chen, Liu, Wang, and Yin]{chen2023on}
Ziang Chen, Jialin Liu, Xinshang Wang, and Wotao Yin.
\newblock On representing linear programs by graph neural networks.
\newblock In \emph{International Conference on Learning Representations}, 2023{\natexlab{a}}.
\newblock URL \url{https://openreview.net/forum?id=cP2QVK-uygd}.

\bibitem[Chen et~al.(2023{\natexlab{b}})Chen, Li, Liu, and Hong]{chen2023characterizing}
Zizhang Chen, Peizhao Li, Hongfu Liu, and Pengyu Hong.
\newblock Characterizing the influence of graph elements.
\newblock In \emph{International Conference on Learning Representations}, 2023{\natexlab{b}}.
\newblock URL \url{https://openreview.net/forum?id=51GXyzOKOp}.

\bibitem[Chiang et~al.(2019)Chiang, Liu, Si, Li, Bengio, and Hsieh]{chiang2019cluster}
Wei-Lin Chiang, Xuanqing Liu, Si~Si, Yang Li, Samy Bengio, and Cho-Jui Hsieh.
\newblock Cluster-gcn: An efficient algorithm for training deep and large graph convolutional networks.
\newblock In \emph{Proceedings of the 25th ACM SIGKDD international conference on knowledge discovery \& data mining}, pp.\  257--266, 2019.

\bibitem[Chidambaram et~al.(2022)Chidambaram, Wang, Hu, Wu, and Ge]{chidambaram2022towards}
Muthu Chidambaram, Xiang Wang, Yuzheng Hu, Chenwei Wu, and Rong Ge.
\newblock Towards understanding the data dependency of mixup-style training.
\newblock In \emph{International Conference on Learning Representations}, 2022.
\newblock URL \url{https://openreview.net/forum?id=ieNJYujcGDO}.

\bibitem[Chun et~al.(2020)Chun, Oh, Yun, Han, Choe, and Yoo]{chun2020empirical}
Sanghyuk Chun, Seong~Joon Oh, Sangdoo Yun, Dongyoon Han, Junsuk Choe, and Youngjoon Yoo.
\newblock An empirical evaluation on robustness and uncertainty of regularization methods.
\newblock \emph{arXiv preprint arXiv:2003.03879}, 2020.

\bibitem[Cong et~al.(2023)Cong, Zhang, Kang, Yuan, Wu, Zhou, Tong, and Mahdavi]{cong2023do}
Weilin Cong, Si~Zhang, Jian Kang, Baichuan Yuan, Hao Wu, Xin Zhou, Hanghang Tong, and Mehrdad Mahdavi.
\newblock Do we really need complicated model architectures for temporal networks?
\newblock In \emph{International Conference on Learning Representations}, 2023.
\newblock URL \url{https://openreview.net/forum?id=ayPPc0SyLv1}.

\bibitem[Corso et~al.(2023)Corso, St{\"a}rk, Jing, Barzilay, and Jaakkola]{corso2023diffdock}
Gabriele Corso, Hannes St{\"a}rk, Bowen Jing, Regina Barzilay, and Tommi~S. Jaakkola.
\newblock Diffdock: Diffusion steps, twists, and turns for molecular docking.
\newblock In \emph{The Eleventh International Conference on Learning Representations}, 2023.
\newblock URL \url{https://openreview.net/forum?id=kKF8_K-mBbS}.

\bibitem[Deng et~al.(2022)Deng, Lian, Wu, and Chen]{deng2022graph}
Leyan Deng, Defu Lian, Chenwang Wu, and Enhong Chen.
\newblock Graph convolution network based recommender systems: Learning guarantee and item mixture powered strategy.
\newblock In Alice~H. Oh, Alekh Agarwal, Danielle Belgrave, and Kyunghyun Cho (eds.), \emph{Advances in Neural Information Processing Systems}, 2022.
\newblock URL \url{https://openreview.net/forum?id=aUoCgjJfmY9}.

\bibitem[Di~Giovanni et~al.(2022)Di~Giovanni, Rowbottom, Chamberlain, Markovich, and Bronstein]{di2022graph}
Francesco Di~Giovanni, James Rowbottom, Benjamin~P Chamberlain, Thomas Markovich, and Michael~M Bronstein.
\newblock Graph neural networks as gradient flows.
\newblock \emph{arXiv preprint arXiv:2206.10991}, 2022.

\bibitem[Du et~al.(2019)Du, Hou, Salakhutdinov, Poczos, Wang, and Xu]{du2019graph}
Simon~S Du, Kangcheng Hou, Russ~R Salakhutdinov, Barnabas Poczos, Ruosong Wang, and Keyulu Xu.
\newblock Graph neural tangent kernel: Fusing graph neural networks with graph kernels.
\newblock \emph{Advances in neural information processing systems}, 32, 2019.

\bibitem[Dudzik \& Veli{\v{c}}kovi{\'c}(2022)Dudzik and Veli{\v{c}}kovi{\'c}]{dudzik2022graph}
Andrew Dudzik and Petar Veli{\v{c}}kovi{\'c}.
\newblock Graph neural networks are dynamic programmers.
\newblock \emph{arXiv preprint arXiv:2203.15544}, 2022.

\bibitem[Fan et~al.(2019)Fan, Ma, Li, He, Zhao, Tang, and Yin]{fan2019graph}
Wenqi Fan, Yao Ma, Qing Li, Yuan He, Eric Zhao, Jiliang Tang, and Dawei Yin.
\newblock Graph neural networks for social recommendation.
\newblock In \emph{The world wide web conference}, pp.\  417--426, 2019.

\bibitem[Frasca et~al.(2020)Frasca, Rossi, Eynard, Chamberlain, Bronstein, and Monti]{frasca2020sign}
Fabrizio Frasca, Emanuele Rossi, Davide Eynard, Ben Chamberlain, Michael Bronstein, and Federico Monti.
\newblock Sign: Scalable inception graph neural networks.
\newblock \emph{arXiv preprint arXiv:2004.11198}, 2020.

\bibitem[Garg et~al.(2020)Garg, Jegelka, and Jaakkola]{garg2020generalization}
Vikas Garg, Stefanie Jegelka, and Tommi Jaakkola.
\newblock Generalization and representational limits of graph neural networks.
\newblock In \emph{International Conference on Machine Learning}, pp.\  3419--3430. PMLR, 2020.

\bibitem[Gasteiger et~al.(2018)Gasteiger, Bojchevski, and G{\"u}nnemann]{gasteiger2018predict}
Johannes Gasteiger, Aleksandar Bojchevski, and Stephan G{\"u}nnemann.
\newblock Predict then propagate: Graph neural networks meet personalized pagerank.
\newblock In \emph{International Conference on Learning Representations}, 2018.

\bibitem[Guo et~al.(2019)Guo, Mao, and Zhang]{guo2019mixup}
Hongyu Guo, Yongyi Mao, and Richong Zhang.
\newblock Mixup as locally linear out-of-manifold regularization.
\newblock In \emph{Proceedings of the AAAI Conference on Artificial Intelligence}, volume~33, pp.\  3714--3722, 2019.

\bibitem[Hamilton et~al.(2017)Hamilton, Ying, and Leskovec]{hamilton2017inductive}
Will Hamilton, Zhitao Ying, and Jure Leskovec.
\newblock Inductive representation learning on large graphs.
\newblock In \emph{NeurIPS}, pp.\  1024--1034, 2017.

\bibitem[Han et~al.(2022{\natexlab{a}})Han, Jiang, Liu, and Hu]{han2022g}
Xiaotian Han, Zhimeng Jiang, Ninghao Liu, and Xia Hu.
\newblock G-mixup: Graph data augmentation for graph classification.
\newblock In \emph{International Conference on Machine Learning}, pp.\  8230--8248. PMLR, 2022{\natexlab{a}}.

\bibitem[Han et~al.(2022{\natexlab{b}})Han, Zhao, Liu, Hu, and Shah]{han2022mlpinit}
Xiaotian Han, Tong Zhao, Yozen Liu, Xia Hu, and Neil Shah.
\newblock Mlpinit: Embarrassingly simple gnn training acceleration with mlp initialization.
\newblock \emph{arXiv preprint arXiv:2210.00102}, 2022{\natexlab{b}}.

\bibitem[Hladi{\v{s}} et~al.(2023)Hladi{\v{s}}, Lalis, Fiorucci, and Topin]{hladi2023matching}
Matej Hladi{\v{s}}, Maxence Lalis, Sebastien Fiorucci, and J{\'e}r{\'e}mie Topin.
\newblock Matching receptor to odorant with protein language and graph neural networks.
\newblock In \emph{International Conference on Learning Representations}, 2023.
\newblock URL \url{https://openreview.net/forum?id=q9VherQJd8_}.

\bibitem[Hu et~al.(2021)Hu, You, Wang, Wang, Zhou, and Gao]{hu2021graph}
Yang Hu, Haoxuan You, Zhecan Wang, Zhicheng Wang, Erjin Zhou, and Yue Gao.
\newblock Graph-mlp: node classification without message passing in graph.
\newblock \emph{arXiv preprint arXiv:2106.04051}, 2021.

\bibitem[Huang et~al.(2020)Huang, He, Singh, Lim, and Benson]{huang2020combining}
Qian Huang, Horace He, Abhay Singh, Ser-Nam Lim, and Austin~R Benson.
\newblock Combining label propagation and simple models out-performs graph neural networks.
\newblock \emph{arXiv preprint arXiv:2010.13993}, 2020.

\bibitem[Hui et~al.(2023)Hui, Yan, Ma, and Ku]{hui2023rethinking}
Bo~Hui, Da~Yan, Xiaolong Ma, and Wei-Shinn Ku.
\newblock Rethinking graph lottery tickets: Graph sparsity matters.
\newblock In \emph{International Conference on Learning Representations}, 2023.
\newblock URL \url{https://openreview.net/forum?id=fjh7UGQgOB}.

\bibitem[Kim et~al.(2020)Kim, Choo, and Song]{kim2020puzzle}
Jang-Hyun Kim, Wonho Choo, and Hyun~Oh Song.
\newblock Puzzle mix: Exploiting saliency and local statistics for optimal mixup.
\newblock In \emph{International Conference on Machine Learning}, pp.\  5275--5285. PMLR, 2020.

\bibitem[Kim et~al.(2021)Kim, Choo, Jeong, and Song]{kim2021co}
Jang-Hyun Kim, Wonho Choo, Hosan Jeong, and Hyun~Oh Song.
\newblock Co-mixup: Saliency guided joint mixup with supermodular diversity.
\newblock \emph{arXiv preprint arXiv:2102.03065}, 2021.

\bibitem[Kingma \& Ba(2015)Kingma and Ba]{kingma2015adam}
Diederik~P Kingma and Jimmy Ba.
\newblock Adam: A method for stochastic optimization.
\newblock In \emph{ICLR}, 2015.

\bibitem[Kipf \& Welling(2016{\natexlab{a}})Kipf and Welling]{kipf2016semi}
Thomas~N Kipf and Max Welling.
\newblock Semi-supervised classification with graph convolutional networks.
\newblock \emph{arXiv preprint arXiv:1609.02907}, 2016{\natexlab{a}}.

\bibitem[Kipf \& Welling(2016{\natexlab{b}})Kipf and Welling]{kipf2016variational}
Thomas~N Kipf and Max Welling.
\newblock Variational graph auto-encoders.
\newblock \emph{arXiv preprint arXiv:1611.07308}, 2016{\natexlab{b}}.

\bibitem[Lamb et~al.(2019)Lamb, Verma, Kannala, and Bengio]{lamb2019interpolated}
Alex Lamb, Vikas Verma, Juho Kannala, and Yoshua Bengio.
\newblock Interpolated adversarial training: Achieving robust neural networks without sacrificing too much accuracy.
\newblock In \emph{Proceedings of the 12th ACM Workshop on Artificial Intelligence and Security}, pp.\  95--103, 2019.

\bibitem[Liao \& Smidt(2023)Liao and Smidt]{liao2023equiformer}
Yi-Lun Liao and Tess Smidt.
\newblock Equiformer: Equivariant graph attention transformer for 3d atomistic graphs.
\newblock In \emph{International Conference on Learning Representations}, 2023.
\newblock URL \url{https://openreview.net/forum?id=KwmPfARgOTD}.

\bibitem[Liu et~al.(2022)Liu, Zhou, Yang, Li, Chen, and Hu]{liu2022exact}
Zirui Liu, Kaixiong Zhou, Fan Yang, Li~Li, Rui Chen, and Xia Hu.
\newblock {EXACT}: Scalable graph neural networks training via extreme activation compression.
\newblock In \emph{International Conference on Learning Representations}, 2022.
\newblock URL \url{https://openreview.net/forum?id=vkaMaq95_rX}.

\bibitem[Lopez-Paz et~al.(2023)Lopez-Paz, Belghazi, Bouchacourt, Dohmatob, Idrissi, Sagun, and Saxe]{lopezpaz2023an}
David Lopez-Paz, Ishmael Belghazi, Diane Bouchacourt, Elvis Dohmatob, Badr~Youbi Idrissi, Levent Sagun, and Andrew~M Saxe.
\newblock An ensemble view on mixup, 2023.
\newblock URL \url{https://openreview.net/forum?id=k_iNqflnekU}.

\bibitem[Nt \& Maehara(2019)Nt and Maehara]{nt2019revisiting}
Hoang Nt and Takanori Maehara.
\newblock Revisiting graph neural networks: All we have is low-pass filters.
\newblock \emph{arXiv preprint arXiv:1905.09550}, 2019.

\bibitem[Pang et~al.(2019)Pang, Xu, and Zhu]{pang2019mixup}
Tianyu Pang, Kun Xu, and Jun Zhu.
\newblock Mixup inference: Better exploiting mixup to defend adversarial attacks.
\newblock \emph{arXiv preprint arXiv:1909.11515}, 2019.

\bibitem[Sabanayagam et~al.(2022)Sabanayagam, Esser, and Ghoshdastidar]{sabanayagam2022representation}
Mahalakshmi Sabanayagam, Pascal Esser, and Debarghya Ghoshdastidar.
\newblock Representation power of graph convolutions: Neural tangent kernel analysis.
\newblock \emph{arXiv preprint arXiv:2210.09809}, 2022.

\bibitem[Shi et~al.(2023)Shi, Liang, and Wang]{shi2023lmc}
Zhihao Shi, Xize Liang, and Jie Wang.
\newblock {LMC}: Fast training of {GNN}s via subgraph sampling with provable convergence.
\newblock In \emph{The Eleventh International Conference on Learning Representations}, 2023.
\newblock URL \url{https://openreview.net/forum?id=5VBBA91N6n}.

\bibitem[Sun et~al.(2021)Sun, Gu, and Hu]{sun2021scalable}
Chuxiong Sun, Hongming Gu, and Jie Hu.
\newblock Scalable and adaptive graph neural networks with self-label-enhanced training.
\newblock \emph{arXiv preprint arXiv:2104.09376}, 2021.

\bibitem[Sun et~al.(2022)Sun, Dai, and Yu]{sun2022does}
Ruoxi Sun, Hanjun Dai, and Adams~Wei Yu.
\newblock Does {GNN} pretraining help molecular representation?
\newblock In Alice~H. Oh, Alekh Agarwal, Danielle Belgrave, and Kyunghyun Cho (eds.), \emph{Advances in Neural Information Processing Systems}, 2022.
\newblock URL \url{https://openreview.net/forum?id=uytgM9N0vlR}.

\bibitem[Tang et~al.(2022)Tang, Liu, He, Wang, and Shah]{tang2022friend}
Xianfeng Tang, Yozen Liu, Xinran He, Suhang Wang, and Neil Shah.
\newblock Friend story ranking with edge-contextual local graph convolutions.
\newblock In \emph{Proceedings of the Fifteenth ACM International Conference on Web Search and Data Mining}, pp.\  1007--1015, 2022.

\bibitem[Thulasidasan et~al.(2019)Thulasidasan, Chennupati, Bilmes, Bhattacharya, and Michalak]{thulasidasan2019mixup}
Sunil Thulasidasan, Gopinath Chennupati, Jeff~A Bilmes, Tanmoy Bhattacharya, and Sarah Michalak.
\newblock On mixup training: Improved calibration and predictive uncertainty for deep neural networks.
\newblock \emph{Advances in Neural Information Processing Systems}, 32, 2019.

\bibitem[Tian et~al.(2023)Tian, Zhang, Guo, Zhang, and Chawla]{tian2023learning}
Yijun Tian, Chuxu Zhang, Zhichun Guo, Xiangliang Zhang, and Nitesh Chawla.
\newblock Learning {MLP}s on graphs: A unified view of effectiveness, robustness, and efficiency.
\newblock In \emph{International Conference on Learning Representations}, 2023.
\newblock URL \url{https://openreview.net/forum?id=Cs3r5KLdoj}.

\bibitem[Van~der Maaten \& Hinton(2008)Van~der Maaten and Hinton]{van2008visualizing}
Laurens Van~der Maaten and Geoffrey Hinton.
\newblock Visualizing data using t-sne.
\newblock \emph{Journal of machine learning research}, 9\penalty0 (11), 2008.

\bibitem[Veli{\v{c}}kovi{\'c} et~al.(2018)Veli{\v{c}}kovi{\'c}, Cucurull, Casanova, Romero, Li{\`o}, and Bengio]{velivckovic2018graph}
Petar Veli{\v{c}}kovi{\'c}, Guillem Cucurull, Arantxa Casanova, Adriana Romero, Pietro Li{\`o}, and Yoshua Bengio.
\newblock Graph attention networks.
\newblock In \emph{International Conference on Learning Representations}, 2018.

\bibitem[Verma et~al.(2019)Verma, Lamb, Beckham, Najafi, Mitliagkas, Lopez-Paz, and Bengio]{verma2019manifold}
Vikas Verma, Alex Lamb, Christopher Beckham, Amir Najafi, Ioannis Mitliagkas, David Lopez-Paz, and Yoshua Bengio.
\newblock Manifold mixup: Better representations by interpolating hidden states.
\newblock In \emph{International Conference on Machine Learning}, pp.\  6438--6447. PMLR, 2019.

\bibitem[Wang et~al.(2023{\natexlab{a}})Wang, Liang, Wang, Wang, Gu, Fang, and Wang]{wang2023searching}
Kun Wang, Yuxuan Liang, Pengkun Wang, Xu~Wang, Pengfei Gu, Junfeng Fang, and Yang Wang.
\newblock Searching lottery tickets in graph neural networks: A dual perspective.
\newblock In \emph{The Eleventh International Conference on Learning Representations}, 2023{\natexlab{a}}.
\newblock URL \url{https://openreview.net/forum?id=Dvs-a3aymPe}.

\bibitem[Wang et~al.(2021)Wang, Wang, Liang, Cai, and Hooi]{wang2021mixup}
Yiwei Wang, Wei Wang, Yuxuan Liang, Yujun Cai, and Bryan Hooi.
\newblock Mixup for node and graph classification.
\newblock In \emph{Proceedings of the Web Conference 2021}, pp.\  3663--3674, 2021.

\bibitem[Wang et~al.(2023{\natexlab{b}})Wang, Nie, Qiao, Xiao, Baraniuk, and Anandkumar]{wang2023retrievalbased}
Zichao Wang, Weili Nie, Zhuoran Qiao, Chaowei Xiao, Richard Baraniuk, and Anima Anandkumar.
\newblock Retrieval-based controllable molecule generation.
\newblock In \emph{International Conference on Learning Representations}, 2023{\natexlab{b}}.
\newblock URL \url{https://openreview.net/forum?id=vDFA1tpuLvk}.

\bibitem[Wei et~al.(2022)Wei, Liang, Liu, and Wang]{wei2022contrastive}
Chunyu Wei, Jian Liang, Di~Liu, and Fei Wang.
\newblock Contrastive graph structure learning via information bottleneck for recommendation.
\newblock In Alice~H. Oh, Alekh Agarwal, Danielle Belgrave, and Kyunghyun Cho (eds.), \emph{Advances in Neural Information Processing Systems}, 2022.
\newblock URL \url{https://openreview.net/forum?id=lhl_rYNdiH6}.

\bibitem[Wu et~al.(2019)Wu, Souza, Zhang, Fifty, Yu, and Weinberger]{wu2019simplifying}
Felix Wu, Amauri Souza, Tianyi Zhang, Christopher Fifty, Tao Yu, and Kilian Weinberger.
\newblock Simplifying graph convolutional networks.
\newblock In \emph{International conference on machine learning}, pp.\  6861--6871. PMLR, 2019.

\bibitem[Wu et~al.(2020)Wu, Pan, Chen, Long, Zhang, and Philip]{wu2020comprehensive}
Zonghan Wu, Shirui Pan, Fengwen Chen, Guodong Long, Chengqi Zhang, and S~Yu Philip.
\newblock A comprehensive survey on graph neural networks.
\newblock \emph{IEEE transactions on neural networks and learning systems}, 32\penalty0 (1):\penalty0 4--24, 2020.

\bibitem[Xia et~al.(2023)Xia, Zhao, Hu, Gao, Tan, Liu, Li, and Li]{xia2023molebert}
Jun Xia, Chengshuai Zhao, Bozhen Hu, Zhangyang Gao, Cheng Tan, Yue Liu, Siyuan Li, and Stan~Z. Li.
\newblock Mole-{BERT}: Rethinking pre-training graph neural networks for molecules.
\newblock In \emph{The Eleventh International Conference on Learning Representations}, 2023.
\newblock URL \url{https://openreview.net/forum?id=jevY-DtiZTR}.

\bibitem[Xu et~al.(2018)Xu, Li, Tian, Sonobe, Kawarabayashi, and Jegelka]{xu2018representation}
Keyulu Xu, Chengtao Li, Yonglong Tian, Tomohiro Sonobe, Ken-ichi Kawarabayashi, and Stefanie Jegelka.
\newblock Representation learning on graphs with jumping knowledge networks.
\newblock In \emph{International conference on machine learning}, pp.\  5453--5462. PMLR, 2018.

\bibitem[Yang et~al.(2023)Yang, Wu, Wang, and Yan]{yang2022graph}
Chenxiao Yang, Qitian Wu, Jiahua Wang, and Junchi Yan.
\newblock Graph neural networks are inherently good generalizers: Insights by bridging gnns and mlps.
\newblock \emph{International Conference on Learning Representations}, 2023.

\bibitem[You et~al.(2020)You, Ying, and Leskovec]{you2020design}
Jiaxuan You, Zhitao Ying, and Jure Leskovec.
\newblock Design space for graph neural networks.
\newblock \emph{Advances in Neural Information Processing Systems}, 33:\penalty0 17009--17021, 2020.

\bibitem[Yun et~al.(2019)Yun, Han, Oh, Chun, Choe, and Yoo]{yun2019cutmix}
Sangdoo Yun, Dongyoon Han, Seong~Joon Oh, Sanghyuk Chun, Junsuk Choe, and Youngjoon Yoo.
\newblock Cutmix: Regularization strategy to train strong classifiers with localizable features.
\newblock In \emph{Proceedings of the IEEE/CVF international conference on computer vision}, pp.\  6023--6032, 2019.

\bibitem[Zeng et~al.(2019)Zeng, Zhou, Srivastava, Kannan, and Prasanna]{zeng2019graphsaint}
Hanqing Zeng, Hongkuan Zhou, Ajitesh Srivastava, Rajgopal Kannan, and Viktor Prasanna.
\newblock Graphsaint: Graph sampling based inductive learning method.
\newblock In \emph{International Conference on Learning Representations}, 2019.

\bibitem[Zhang et~al.(2023{\natexlab{a}})Zhang, Luo, Wang, and He]{zhang2023rethinking}
Bohang Zhang, Shengjie Luo, Liwei Wang, and Di~He.
\newblock Rethinking the expressive power of {GNN}s via graph biconnectivity.
\newblock In \emph{International Conference on Learning Representations}, 2023{\natexlab{a}}.
\newblock URL \url{https://openreview.net/forum?id=r9hNv76KoT3}.

\bibitem[Zhang et~al.(2018)Zhang, Cisse, Dauphin, and Lopez-Paz]{zhang2018mixup}
Hongyi Zhang, Moustapha Cisse, Yann~N Dauphin, and David Lopez-Paz.
\newblock mixup: Beyond empirical risk minimization.
\newblock In \emph{International Conference on Learning Representations}, 2018.

\bibitem[Zhang et~al.(2020)Zhang, Deng, Kawaguchi, Ghorbani, and Zou]{zhang2020does}
Linjun Zhang, Zhun Deng, Kenji Kawaguchi, Amirata Ghorbani, and James Zou.
\newblock How does mixup help with robustness and generalization?
\newblock In \emph{International Conference on Learning Representations}, 2020.

\bibitem[Zhang et~al.(2022{\natexlab{a}})Zhang, Deng, Kawaguchi, and Zou]{zhang2022and}
Linjun Zhang, Zhun Deng, Kenji Kawaguchi, and James Zou.
\newblock When and how mixup improves calibration.
\newblock In \emph{International Conference on Machine Learning}, pp.\  26135--26160. PMLR, 2022{\natexlab{a}}.

\bibitem[Zhang et~al.(2021)Zhang, Liu, Sun, and Shah]{zhang2021graph}
Shichang Zhang, Yozen Liu, Yizhou Sun, and Neil Shah.
\newblock Graph-less neural networks: Teaching old mlps new tricks via distillation.
\newblock In \emph{International Conference on Learning Representations}, 2021.

\bibitem[Zhang et~al.(2023{\natexlab{b}})Zhang, Wang, Chen, Liu, Lu, and Liu]{zhang2023joint}
Shuai Zhang, Meng Wang, Pin-Yu Chen, Sijia Liu, Songtao Lu, and Miao Liu.
\newblock Joint edge-model sparse learning is provably efficient for graph neural networks.
\newblock In \emph{The Eleventh International Conference on Learning Representations}, 2023{\natexlab{b}}.
\newblock URL \url{https://openreview.net/forum?id=4UldFtZ_CVF}.

\bibitem[Zhang et~al.(2022{\natexlab{b}})Zhang, Yin, Sheng, Li, Ouyang, Li, Tao, Yang, and Cui]{zhang2022graph}
Wentao Zhang, Ziqi Yin, Zeang Sheng, Yang Li, Wen Ouyang, Xiaosen Li, Yangyu Tao, Zhi Yang, and Bin Cui.
\newblock Graph attention multi-layer perceptron.
\newblock \emph{arXiv preprint arXiv:2206.04355}, 2022{\natexlab{b}}.

\bibitem[Zhang et~al.(2023{\natexlab{c}})Zhang, Cai, Shi, and Tang]{zhang2023ebind}
Yangtian Zhang, Huiyu Cai, Chence Shi, and Jian Tang.
\newblock E3bind: An end-to-end equivariant network for protein-ligand docking.
\newblock In \emph{International Conference on Learning Representations}, 2023{\natexlab{c}}.
\newblock URL \url{https://openreview.net/forum?id=sO1QiAftQFv}.

\bibitem[ZHANG et~al.(2023)ZHANG, Liu, Zheng, and Min]{zhang2023molecule}
ZAIXI ZHANG, Qi~Liu, Shuxin Zheng, and Yaosen Min.
\newblock Molecule generation for target protein binding with structural motifs.
\newblock In \emph{International Conference on Learning Representations}, 2023.
\newblock URL \url{https://openreview.net/forum?id=Rq13idF0F73}.

\bibitem[Zhao et~al.(2023)Zhao, Qu, Li, Yan, Liu, Li, Xie, and Tang]{zhao2023learning}
Jianan Zhao, Meng Qu, Chaozhuo Li, Hao Yan, Qian Liu, Rui Li, Xing Xie, and Jian Tang.
\newblock Learning on large-scale text-attributed graphs via variational inference.
\newblock In \emph{International Conference on Learning Representations}, 2023.
\newblock URL \url{https://openreview.net/forum?id=q0nmYciuuZN}.

\bibitem[Zhou et~al.(2020)Zhou, Cui, Hu, Zhang, Yang, Liu, Wang, Li, and Sun]{zhou2020graph}
Jie Zhou, Ganqu Cui, Shengding Hu, Zhengyan Zhang, Cheng Yang, Zhiyuan Liu, Lifeng Wang, Changcheng Li, and Maosong Sun.
\newblock Graph neural networks: A review of methods and applications.
\newblock \emph{AI Open}, 1:\penalty0 57--81, 2020.

\bibitem[Zhu et~al.(2023{\natexlab{a}})Zhu, Wu, Wang, Xia, Xie, Meng, Wu, Qin, Zhou, Li, and Liu]{zhu2023mathcalognn}
Jinhua Zhu, Kehan Wu, Bohan Wang, Yingce Xia, Shufang Xie, Qi~Meng, Lijun Wu, Tao Qin, Wengang Zhou, Houqiang Li, and Tie-Yan Liu.
\newblock \${\textbackslash}mathcal\{O\}\$-{GNN}: incorporating ring priors into molecular modeling.
\newblock In \emph{International Conference on Learning Representations}, 2023{\natexlab{a}}.
\newblock URL \url{https://openreview.net/forum?id=5cFfz6yMVPU}.

\bibitem[Zhu et~al.(2023{\natexlab{b}})Zhu, Li, Mo, Hu, Li, Liu, Liang, and Cheng]{zhu2023rm}
Zeyu Zhu, Fanrong Li, Zitao Mo, Qinghao Hu, Gang Li, Zejian Liu, Xiaoyao Liang, and Jian Cheng.
\newblock \${\textbackslash}rm a{\textasciicircum}2q\$: Aggregation-aware quantization for graph neural networks.
\newblock In \emph{The Eleventh International Conference on Learning Representations}, 2023{\natexlab{b}}.
\newblock URL \url{https://openreview.net/forum?id=7L2mgi0TNEP}.

\end{thebibliography}
\bibliographystyle{tmlr}

\clearpage
\appendix
\part{Appendix} 
\parttoc

\section{Related Works}\label{sec:app:rela}
In this appendix, we discuss more related work to our paper.

\subsection{Graph Neural Networks}
Starting from graph convolutional network~\citep{kipf2016semi}, graph neural networks have been applied to various graph learning tasks and show great power for graph learning tasks. Despite the practical power of graph neural network, understanding the generalization and working mechanism of the graph neural networks are still in its infancy~\citep{garg2020generalization,zhang2023rethinking,yang2022graph,baranwal2023effects}. The previous works try to understand graph neural networks from different perspectives, such as signal processing~\citep{nt2019revisiting,bo2021beyond,bianchi2021graph}, gradient flow~\citep{di2022graph}, dynamic programming\citep{dudzik2022graph}, neural tangent kernels~\citep{yang2022graph,du2019graph,sabanayagam2022representation} and influence function~\citep{chen2023characterizing}. There is also a line of works try to understand graph neural networks by analyzing the connection between GNNs and MLPs~\citep{baranwal2023effects,han2022mlpinit,yang2022graph,tian2023learning}.

In our work, we understand graph neural networks through a fresh perspective, Mixup. We believe this work will inspire further research and lead to the development of new techniques for improving the performance and interpretability of GNNs.

\subsection{Mixup and Its Variants}
Mixup~\citep{zhang2018mixup} and its variants~\citep{verma2019manifold,yun2019cutmix,kim2020puzzle,kim2021co} are important data augmentation methods that are effective in improving the generalization performance of deep neural networks. Mixup is used to understand or improve many machine learning techniques. More specifically, deep neural networks trained with Mixup achieve better generalization~\citep{chun2020empirical,zhang2020does,chidambaram2022towards}, calibration~\citep{thulasidasan2019mixup,zhang2022and}, and adversarial robustness~\citep{pang2019mixup,archambault2019mixup,zhang2020does,lamb2019interpolated}.  As a studies data augmentation method, Mixup can be understood as a regularization term to improve the effectiveness and robustness of neural network~\citep{pang2019mixup,archambault2019mixup,zhang2020does,lamb2019interpolated}. Mixup are also used to explain some other techniques, such as ensmeble~\citep{lopezpaz2023an}. Additionally, Mixup technique is also used to augment graph data for better performance of GNNs~\cite{han2022g,wang2021mixup}. 

In this work, we use Mixup to explain and understand the graph neural network, and based on the Mixup, we propose two kinds of MLP that achieve almost equivalent performance to graph neural networks. Also, our work is different form the works that chance the GNN training with Mixup, while we connect the graph convolution and Mixup.

\subsection{Efficient Graph Neural Networks} Although the graph neural networks are powerful for graph learning tasks, they suffer slow computational problems since they involve either sparse matrix multiplication or neighbors fetching during the inference, resulting in active research on the efficiency of graph neural networks. The recent advances in efficient graph neural networks are as follows. Efficient graph neural network~\citep{wang2023searching,shi2023lmc,hui2023rethinking,zhang2023joint,zhu2023rm} includes the sampling-based method~\citep{hamilton2017inductive,chiang2019cluster,zeng2019graphsaint}, quantization~\citep{liu2022exact}, knowledge distillation~\citep{zhang2021graph,tian2023learning}, and graph sparsification~\citep{cai2020graph}. There is a line of work to try to use MLP to accelerate the GNN training and inference~\citep{zhang2022graph,wu2019simplifying,frasca2020sign,sun2021scalable,huang2020combining,hu2021graph}. Simplifying graph convolutional (SGC)~\citep{wu2019simplifying} ``simplifying'' graph convolutional network by decoupling the neighbor aggregation and feature transformation, the following work such as \citet{you2020design} employ the similar basic idea. \citet{zhang2021graph} leverage knowledge distillation to transfer the knowledge from GNN to MLP. \citet{han2022mlpinit} and \citet{yang2022graph} transfer the weight from MLP to GNN. These kinds of methods use MLP for efficient graph neural networks.

Our proposed methods \hmlp and \tmlp are efficient for training and/or inference, which can be categorized as an efficient graph neural network. Besides,  our proposed method provides a new perspective to understand graph neural networks.

\section{Connection between Mixup and More Graph Convolutions}

\subsection{Discussion on More Graph Convolutions} 
We examine the connection between complex graph convolutions used in various graph neural networks and Mixup. We aim to approximately rewrite complex graph convolutions (like those in graph attention networks) into Mixup forms. 
We present a general Mixup form of graph convolution for node $n_i$ in the graph as follows
\begin{equation}\label{eq:gcn_3}
\begin{split}
    (\Tilde{\mathbf{x}}, \Tilde{\mathbf{y}}) = \left( \sum_{j \in {\color{myblue} \mathcal{N}}_i} {\color{myred}\lambda_j} {\color{mygreen}\mathbf{H}_j}, \sum_{j \in {\color{myblue} \mathcal{N}}_i } {\color{myred}\lambda_j} \mathbf{y}_i\right).
\end{split}
\end{equation}

\begin{table}[!h]
    \centering
    \vspace{-18pt}
    \fontsize{9}{12}\selectfont
    \caption{Examples of the relation between graph convolutions to Mixup at k-layer. }\label{tab:more_graphconv}
    \begin{tabular}{cccccc}
    \toprule
            & {\color{myblue} ($\mathcal{N}_i$)}  &  {\color{myred} ($\lambda_j$)}     & {\color{mygreen} ($\mathbf{H}_j$)} \\ \midrule
    GCN     & $\mathcal{N}_i$                     &   $\mathbf{a}_i$                   &  $\mathbf{H}^k$\\
    SGC     & $\mathcal{N}_i^k$                   &   $\mathbf{a}_i^k$                   &  $\mathbf{X}$\\
    GAT     & $\mathcal{N}_i$                     &   Attention Score                  &  $\mathbf{H}^k$\\
    PPNP    & $\mathcal{N}_i$                     &   $\mathbf{a}_i$                   &  Logit\\
    \bottomrule
    \end{tabular}
\end{table}

\cref{eq:gcn_3} shows that graph neural networks architectures vary depending on: 1) how to {\color{myblue} select neighbors set ($\mathcal{N}_i$)}, 2) how to select {\color{myred} aggregation coefficient ($\lambda_j$)}, and 3) {\color{mygreen}what is mixed up ($\mathbf{H}_j$)}. 
For example, if we set $\mathcal{N}=\mathcal{N}_i^2$, $\lambda=\frac{1}{|\mathcal{N}_i^2|}$, the above equation will be SGC. However, there are a lot of advanced graph neural networks, such as GAT~\citep{velivckovic2018graph}) (attention-based GNN), and PPNP~\citep{gasteiger2018predict} (MLP as the backbone). \cref{tab:more_graphconv} shows the relationship between those graph convolutions to Mixup. To include residual connections in some graph convolutions~\citep{chen2020simple,xu2018representation}, we can expand the neighbors set $\mathcal{N}_i$ of node $n_i$ to include its node representation of the previous time step.

\subsection{Experiments on APPNP}

APPNP essentially performs the neighbor aggregation in the last layer. Thus, TMLP and HMLP, according to APPNP, are equivalent to GCN. We report the performance in the table below. The results show that TMLP and HMLP achieve better performance than APPNP.

\begin{table}[!h]
\centering
\caption{Performance Metrics of Different Methods}\label{tab:APPNP}
\begin{tabular}{llll}
\toprule
\textbf{ MLP } & \textbf{ PPNP } & \textbf{ TMLP } & \textbf{ HMLP }  \\ \midrule
73.57          & 83.80           & 88.26           & 86.42 \\
\bottomrule
\end{tabular}
\end{table}

\subsection{GNN with trainable mixing parameter $\lambda$}
In this experiment, we adopted a method similar to GAT to examine the performance of our proposed method, with the mixing parameter $\lambda$ being learned during the training process. The Softmax (Li et al., 2020) is a learnable aggregation operator that normalizes the features of neighbors based on a learnable temperature term. We report the performance in the table below. The results show that the TMLP performed worse than the MLP. This indicates that the learnable mixing parameter $\lambda$ does not work well for our proposed method

\begin{table}[!h]
\centering
\caption{Performance Metrics of Different Methods}\label{tab:gat}
\begin{tabular}{lll}
\toprule
\textbf{ MLP } & \textbf{ GAT-alike } & \textbf{ TMLP }  \\ \midrule
73.57          & 88.42                & 65.62  \\
\bottomrule
\end{tabular}
\end{table}

\section{Additional Experiments}

This appendix presents more experiment results on \hmlp and \tmlp, including 1) comparison experiments of MLP, GCN, and \hmlp/\tmlp; 2) more experimental results on decision boundary.

\subsection{More Experiments for \hmlp}\label{sec:app:exp:hmlp}
\begin{figure}[!htp]
    \centering
    \includegraphics[width=0.99\textwidth]{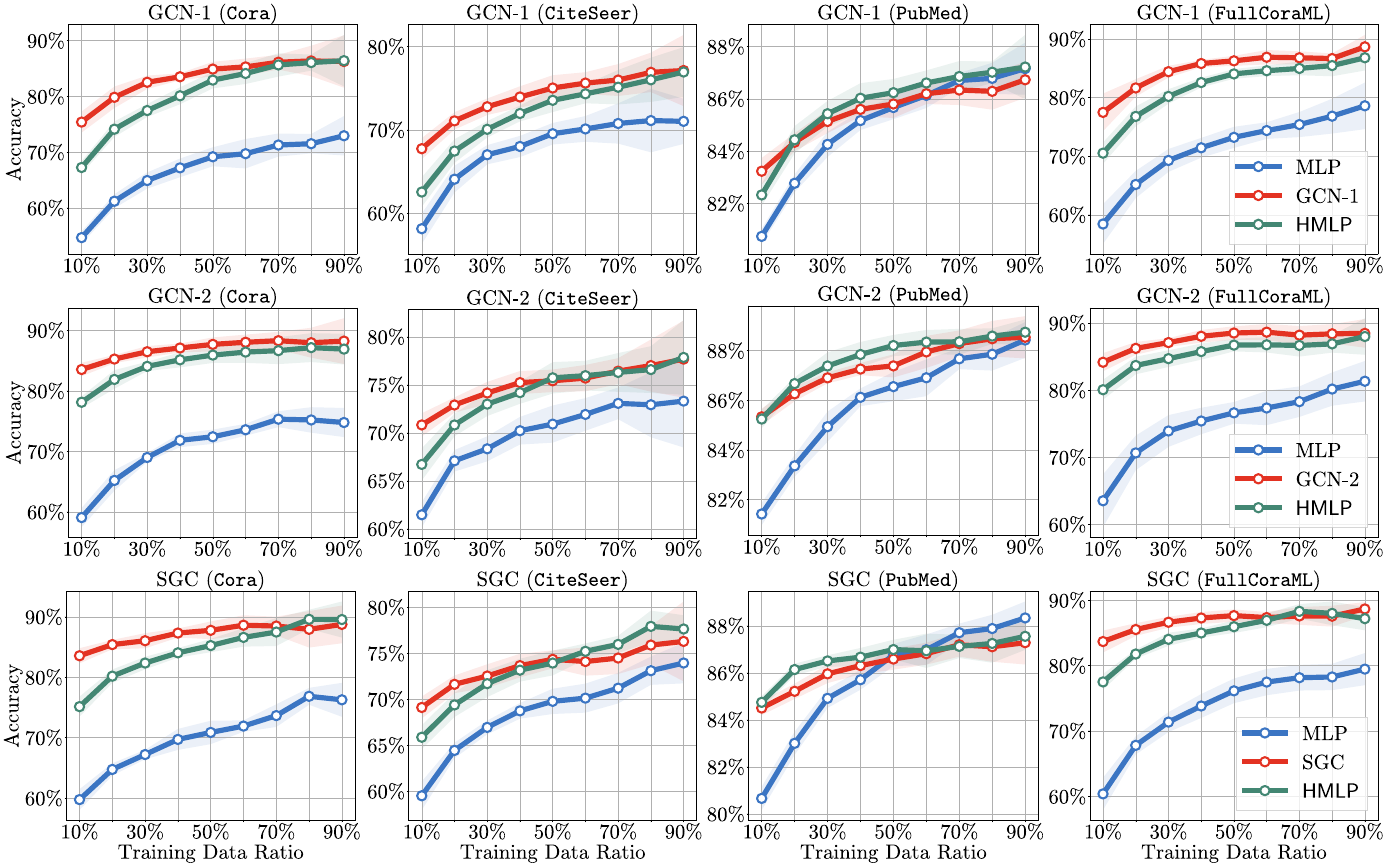}
    \caption{The performance comparison of the {\color{myred}GCN}, {\color{myblue}MLP} and {\color{mygreen}\hmlp} (Ours) with GNN architectures. The x-axis represents the ratio of training data, and the y-axis represents the classification accuracy. The results show that our proposed method ({\color{mygreen}\hmlp}) achieves comparable performance to {\color{myred}GCN}.}\label{fig:pref_hmlp_add}
\end{figure}

We present further experiments comparing MLP, GCN, and \hmlp. These experiments complement those discussed in \cref{fig:pref_hmlp}. We expand on that work by evaluating these methods with more GNN architectures. The results are presented in \cref{fig:pref_hmlp_add}.

The results on more GNN architectures indicate that our proposed \hmlp method not only outperforms traditional MLP, but also attains a performance that is largely on par with GNNs in the majority of instances.

\subsection{More Experiments for \tmlp}\label{sec:app:exp:tmlp}
\begin{figure}[!htp]
    \centering
    \includegraphics[width=0.99\textwidth]{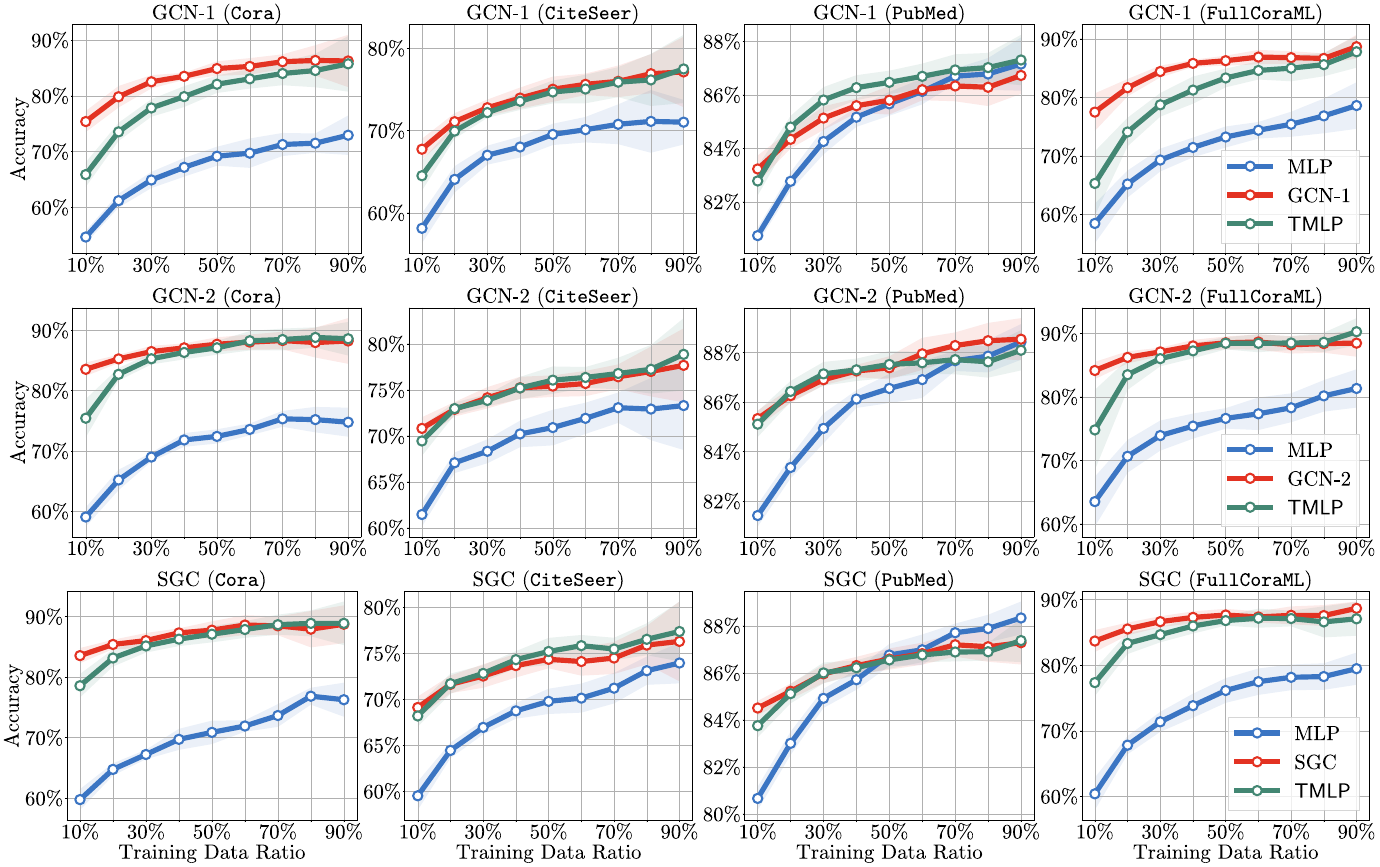}
    \caption{The performance comparison of the {\color{myred}GCNs}, {\color{myblue}MLP} and {\color{mygreen}\tmlp} (Ours). GCN-1 is a one-layer GCN, and GCN-2 is a two-layer GCN. The results show that \tmlp achieves a comparable performance to GCN and SGC. Note that the architecture is essentially MLP in training time and GCN in the test time.}\label{fig:pref_tmlp_add}
\end{figure}

We present further experiments comparing MLP, GCN, and \tmlp. These experiments complement those discussed in \cref{fig:pref_tmlp}. We expand on that work by evaluating these methods with more GNN architectures. The results are presented in \cref{fig:pref_tmlp_add}.

With the evaluation of an expanded selection of GNN architectures, the results reveal that our proposed method \tmlp typically achieves performance comparable to traditional GNNs in most situations.

\subsection{More Experiments on Decision Boundary}\label{sec:app:exp:db}
We provide additional experimental results pertaining to the decision boundary of \tmlp. Our findings indicate that \tmlp tends to shift the node features in the test set away from the pre-learned decision boundary by \tmlp.

\begin{figure}
    \centering
    \includegraphics[width=0.98\textwidth]{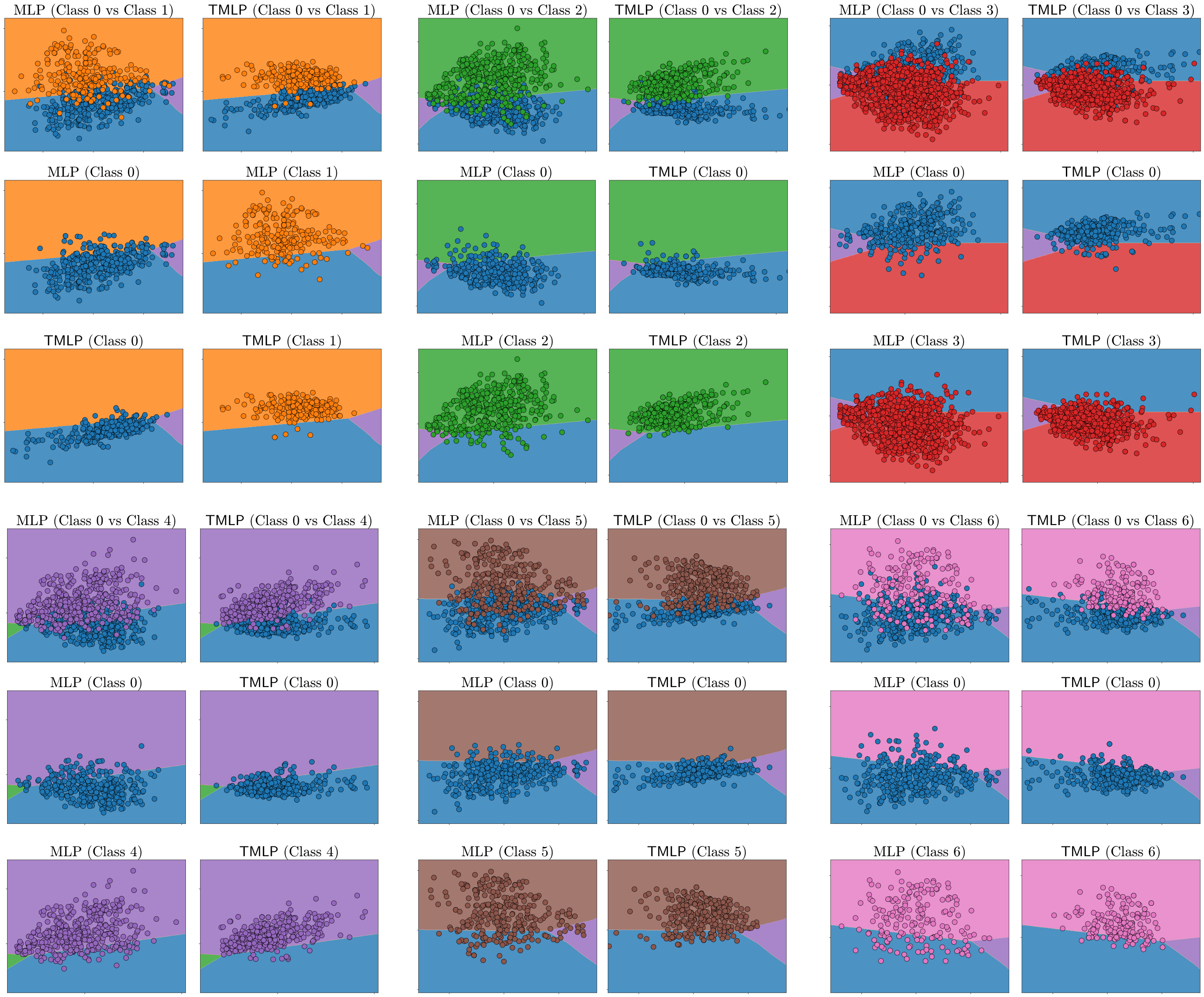}
    \caption{Decision Boundary of MLP and \tmlp. \tmlp can cause node features to move away from the learned boundary.}\label{fig:decision_boundary_all}
\end{figure}

\subsection{More Experiments on Deeper GCNs}\label{sec:app:exp:deeper}
We conducted additional experiments on multi-layer graph neural network (SGC) on the Cora dataset. We use the 20\%\textbackslash 40\%\textbackslash 40\% for train\textbackslash val\textbackslash test data split. The test accuracy was determined based on the results from the validation dataset. The test accuracy is reported in the following table. We observe from the results that our proposed HMLP+TMLP method achieves performance comparable to that of SGC when multiple graph convolutions are utilized.

\begin{table}
\centering
\caption{Performance comparison of GNNs and our proposed method with different layers.}\label{tab:layers}
\begin{tabular}{lcccc}
\toprule
\textbf{ } & \textbf{ 1-layer } & \textbf{ 2-layer } & \textbf{ 3-layer } & \textbf{ 4-layer }  \\ \midrule
MLP        & 69.11              & 68.95              & 68.39              & 69.22               \\
SGC        & 82.38              & 84.50              & 84.23              & 83.95               \\
\hmlp       & 78.56              & 80.41              & 80.96              & 81.15               \\
\tmlp       & 82.10              & 83.76              & 83.49              & 83.67               \\
\hmlp+\tmlp  & 83.12              & 83.95              & 84.87              & 84.41                \\
\bottomrule
\end{tabular}
\end{table}

\subsection{Experiments on More Datasets}
\begin{figure}
    \centering
    \includegraphics[width=1\linewidth]{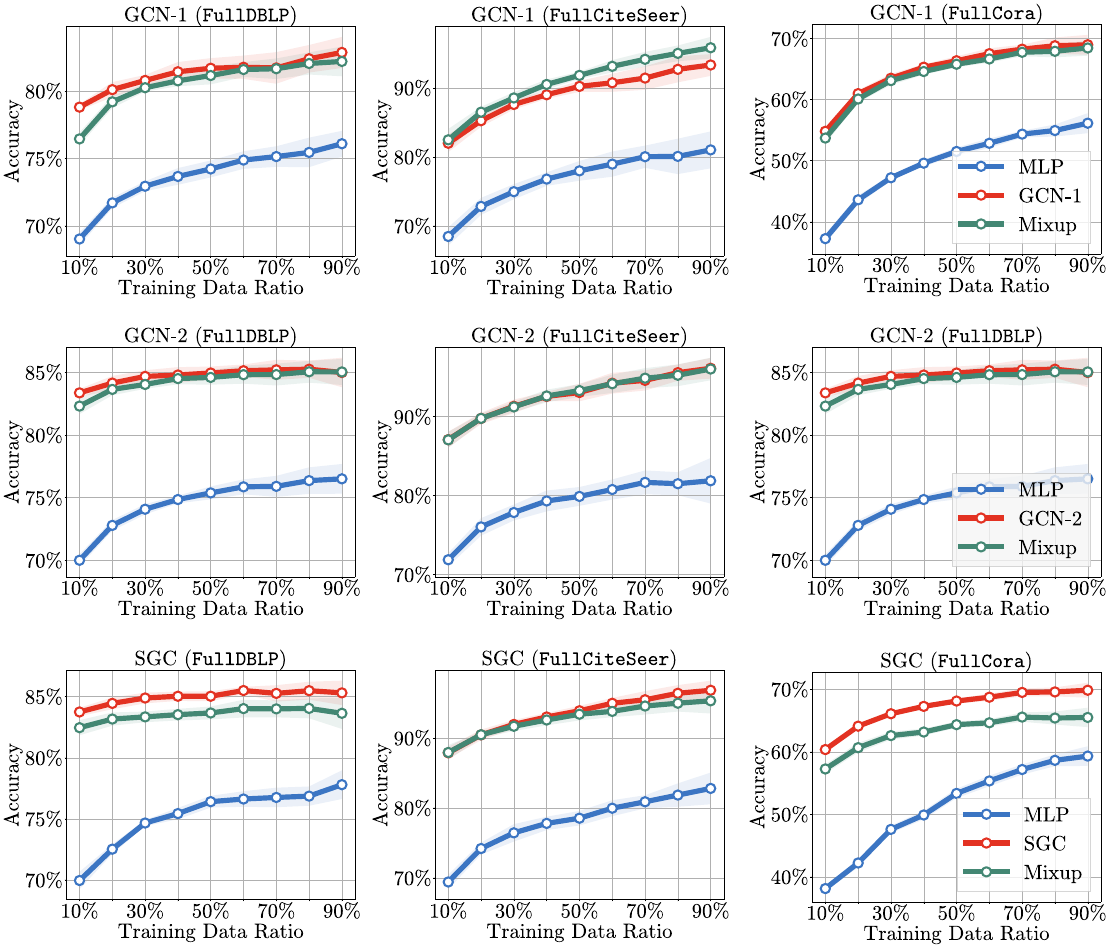}
    \caption{The additional experimental result on unifying HMLP and TMLP. These methods together achieve performance comparable to GNNs on these additional datasets. The additional results show that graph convolution is a mixup under our proposed modifications.}\label{fig:more_data}
\end{figure}

We conducted additional experiments on more datasets, including FullCora, FullSiteSeer, FullDBLP, and the results are presented in \cref{fig:more_data}. The additional results provide more evidence that graph convolution is a mixup under our proposed modifications.

\section{Experiment Setting}\label{sec:app:exp}
In this appendix, we present the running environment for our experiment and the experimental setup for all the experiments.

\subsection{Running Environments}\label{sec:app:exp:runing}

We run all the experiments on NVIDIA RTX A100 on AWS. The code is based on PyTorch~\footnote{\url{https://pytorch.org/}} and torch\_geometric~\footnote{\url{https://pyg.org/}}. The datasets used in this paper are built-in datasets in torch\_geometric, which will be automatically downloaded via the torch\_geometric API.

\subsection{Experiment Setup for \cref{fig:pref_hmlp}}
We provide details of the experimental setup associated with \cref{fig:pref_hmlp}. The primary objective of this experiment was to compare the performance of our proposed \hmlp model with that of the GCN model. We utilized diverse datasets to facilitate a comprehensive evaluation. We set the learning rate to $0.1$ for all methods and datasets, and each was trained for $400$ epochs.Moreover, we did not use weight decay for model training. The test accuracy reported is based on the best results from the validation set. The node classification accuracy served as the evaluation metric for this experiment.

\subsection{Experiment Setup for \cref{fig:pref_tmlp}}
We provide details of the experimental setup associated with \cref{fig:pref_tmlp}. We set the learning rate to $0.1$ for all methods, and each was trained for $400$ epochs. Additionally, weight decay was not utilized in this process. The test accuracy reported is based on the best results from the validation set. The node classification accuracy served as the evaluation metric for this experiment.

\subsection{Experiment Setup for \cref{fig:mlp_all}}
We provide details of the experimental setup associated with \cref{fig:mlp_all}. We set the learning rate to $0.1$ for all methods, and each was trained for $400$ epochs. Additionally, weight decay was not utilized in this process. The test accuracy reported is based on the best results from the validation set. The node classification accuracy served as the evaluation metric for this experiment.

\section{Datasets}\label{sec:app:data}

In this experiment, we use widely adopted node classification benchmarks involving different types of networks: three citation networks (\cora, \citeseer and \pubmed).
The datasets used in this experiment are widely adopted node classification benchmarks. These networks are often used as standard evaluation benchmarks in the field of graph learning.

\begin{itemize}
    \item \cora: is a widely used benchmark data for graph learning, which is a citation network, where each node represents a research paper, and edges denote the citation relation. Node features are extracted from the abstract of the paper.
    
    \item \citeseer is another citation network used frequently in GNN literature. It follows a similar structure to the \cora dataset, where each node represents a scientific document, and the edges are citation information between them. The node features are derived from the text of the documents.
    
    \item \pubmed: PubMed is a citation network composed of scientific papers related to biomedicine. As with the other datasets, nodes represent documents, and edges indicate citation relationships. Node features correspond to term frequency-inverse document frequency (TF-IDF) representations of the documents.
    
    \item \coraml is a subset of the full \cora dataset. It retains the same structure as the \cora dataset but is restricted to documents and citations within the field of machine learning. The node features are the bag-of-words representation for the papers.
\end{itemize}

\end{document}